\documentclass[DTMColor]{ROB-New}

\usepackage[utf8]{inputenc} % allow utf-8 input
\usepackage[T1]{fontenc}    % use 8-bit T1 fonts
\usepackage{hyperref}       % hyperlinks
\usepackage{url}            % simple URL typesetting
\usepackage{booktabs}       % professional-quality tables
\usepackage{amsfonts}       % blackboard math symbols
\usepackage{nicefrac}       % compact symbols for 1/2, etc.
\usepackage{microtype}      % microtypography
\usepackage{amssymb,amsfonts,amsmath,amscd}
\usepackage[printonlyused,withpage]{acronym}
\graphicspath{{figs/}}

\def\acp#1{#1s}
\def\ac#1{#1}
\newcommand{\bbm}{\begin{bmatrix}}
\newcommand{\ebm}{\end{bmatrix}}
\newcommand{\mbf}{\mathbf}
\newcommand{\mbs}[1]{{\boldsymbol{#1}}}
\newcommand{\beq}{\begin{equation}}
\newcommand{\eeq}{\end{equation}}
\newcommand{\bdis}{\begin{displaymath}}
\newcommand{\edis}{\end{displaymath}}
\newcommand{\beqn}[1]{\begin{subequations}\label{eq:#1}\begin{eqnarray}}
\newcommand{\eeqn}{\end{eqnarray}\end{subequations}}

\newcommand{\pri}[1]{\check{#1}}
\newcommand{\mbc}[1]{{\mathcal{#1}}}
\newcommand{\ip}[2]{{\left< #1,#2 \right>}}

\renewcommand{\vec}{\mbox{vec}}
\newcommand{\vech}{\mbox{vech}}

\newcommand{\tr}{\mbox{tr}}

\newcommand{\norm}[1]{\,\downarrow\!{#1}}

\newboolean{narrow}
\setboolean{narrow}{true}

\begin{document}

\authormark{Cambridge Authors}

\articletype{RESEARCH ARTICLE}

\jnlPage{1}{00}
\jyear{2021}
\jdoi{10.1017/xxxxx}

\title{Variational Inference as Iterative Projection in a Bayesian Hilbert Space with Application to Robotic State Estimation}

\author[1]{Timothy D. Barfoot\hyperlink{corr}{*}}
\author[1]{Gabriele M. T. D'Eleuterio}
\address[1]{University of Toronto Institute for Aerospace Studies, 4925 Dufferin Street, Toronto, Ontario, Canada M3H 5T6}
\address{\hypertarget{corr}{*}Corresponding author. \email{tim.barfoot@utoronto.ca}}

\received{xx xxx xxx}
\revised{xx xxx xxx}
\accepted{xx xxx xxx}


\abstract{Variational Bayesian inference is an important machine-learning tool that finds application from statistics to robotics.  The goal is to find an approximate probability density function (\ac{PDF}) from a chosen family that is in some sense `closest' to the full Bayesian posterior.  Closeness is typically defined through the selection of an appropriate loss functional such as the Kullback-Leibler (\ac{KL}) divergence.  In this paper, we explore a new formulation of variational inference by exploiting the fact that (most) PDFs are members of a Bayesian Hilbert space under careful definitions of vector addition, scalar multiplication and an inner product.  We show that, under the right conditions, variational inference based on \ac{KL} divergence can amount to iterative projection, in the Euclidean sense, of the Bayesian posterior onto a subspace corresponding to the selected approximation family.  We work through the details of this general framework for the specific case of the Gaussian approximation family and show the equivalence to another Gaussian variational inference approach.  We furthermore discuss the implications for systems that exhibit sparsity, which is handled naturally in Bayesian space, and give an example of a high-dimensional robotic state estimation problem that can be handled as a result.  We provide some preliminary examples of how the approach could be applied to non-Gaussian inference and discuss the limitations of the approach in detail to encourage follow-on work along these lines.}

\maketitle

%!TEX root =  robotica.tex
%%%%%%%%%%%%%%%%%%%%%%%%%%%%%%%%%%%%%%%%%%%%%%%%%%%%%%%%%%%%%%%%%%%%%%%%%%%%%%%%%%%%%%%%%%%%%%%%%%%
\section{Introduction}
%%%%%%%%%%%%%%%%%%%%%%%%%%%%%%%%%%%%%%%%%%%%%%%%%%%%%%%%%%%%%%%%%%%%%%%%%%%%%%%%%%%%%%%%%%%%%%%%%%%

In 1763, Richard Price published on behalf of his recently deceased friend, the Reverend Thomas Bayes, a paper that introduced what would become the atomic element of probabilistic inference:  Bayes' rule \citep{bayes}.  The paper though was widely ignored.  About a decade later, the same rule was discovered by Pierre-Simon Laplace and, while Laplace laid its foundations, the theory of inference based on this rule became known as Bayesian inference.  So confident was Laplace in the theory that he famously calculated the odds at 11,000 to 1 that the mass of Saturn as determined by a former student was correct to within 1\%, 1,000,000-to-1 odds on the mass of Jupiter \citep[translated from 1825 French edition, pp.~46-47]{Laplace1995}.  (Based on the most recent available data, he would have collected on the bet on Saturn.)   Bayesian inference has been used in a great variety of applications from Henri Poincar\'e's defense of Captain Dreyfus to Alan Turing's breaking of the Enigma code \citep{McGrayne2011}.  In modern day, it provides the crucial framework for inference in such fields as statistics, decision theory, computational neuroscience, machine learning, computer vision, state estimation and robotics.

The objective common to all these applications is the determination of a {\em posterior} probability to test some hypothesis or to calculate some estimate based on {\em prior} information and observed measurements.  However, it is not always possible to find the posterior exactly.  Indeed, we must often resort to approximate techniques.  One such technique, which will occupy us here, is that of {\em variational inference} or {\em variational Bayes} \citep{bishop06}.  In this variational approach, the goal is to find the probability density function that comes closest to the posterior as determined by minimizing a loss functional subject to the constraint that the distribution sought be drawn from a tractable class of densities, for example, where the posterior has to take the form of a Gaussian distribution.  A common choice for the loss functional is the Kullback-Leibler divergence \citep{Csiszar1975, Hinton&vanCamp1993, Jordan&al1999, bishop06, Barber2012, Amari2016, Blei&al2017} although others such as Bregman \citep{Adamcik2014, Painsky&Wornell_arxiv2020}, Wasserstein \citep{Ambrogioni&al2018} and R\'enyi divergences \citep{Li&Turner2016} have been used.

The field of variational inference based on the \ac{KL} divergence is already well trodden although the research is hardly exhausted.  The chosen class of densities from which the approximate posterior is to be shaped is key to variational inference.  In the mean-field approximation, for example, the solution to the minimization of the divergence is constructed as a product of densities from a chosen family of admissible functions such as a Bayesian mixture of Gaussians \citep{bishop06}.  Another possibility is using Bayesian mixtures of exponential families \citep{Wainwright&Jordan2008,Amari2016}.  A number of algorithms by which to execute the minimization exist including the variational EM algorithm, natural gradient descent and Gaussian variational inference.

\begin{figure}[t]
\centering
\includegraphics[width=\textwidth]{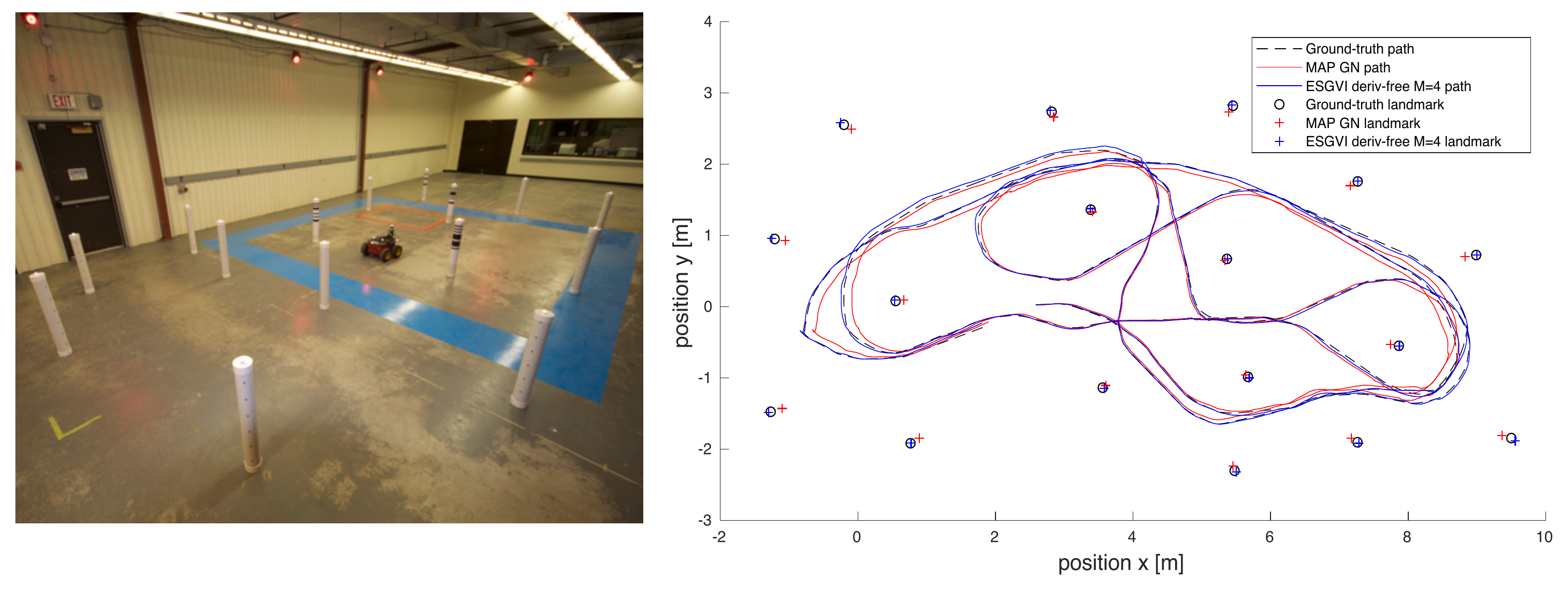}
\caption{A special case of the approach presented in this paper was previously demonstrated \citep{barfoot_ijrr20} to be a useful and practical tool for robotic state estimation.  The method, called Exactly Sparse Gaussian Variational Inference (ESGVI), was used to solve the Simultaneous Localization and Mapping (SLAM) problem, outperforming the standard Maximum A Posteriori (MAP) estimation in certain cases.  The current paper reinterprets this earlier work in a new mathematical formalism.  Figure reproduced from \citep{barfoot_ijrr20}}
\label{fig:robot}
\vspace*{-0.25in}
\end{figure}

\citet{Jordan&al1999} observed that ``there is not as yet a systematic algebra that allows particular variational transformations to be matched optimally to particular graphical models.'' While this was written two decades ago and specifically about graphical models, the remark finds resonance in the present work.

In previous work \citep{barfoot_ijrr20}, we developed a practical robotic state estimation tool based on variational inference and compared it to Maximum A Posteriori (MAP), showing some advantages in certain situations.  For example, the method we developed, dubbed Exactly Space Gaussian Variational Inference (ESGVI), can be used to solve the famous Simultaneous Localization and Mapping (SLAM) problem.  The current paper shows this existing method can be viewed through a different lens, that of iterative projections in a special space known as a {\em Bayesian Hilbert space} or {\em Bayes space} for short \citep{vandenBoogaart2014}.  The primary contribution of this paper is therefore to make this connection between two quite different fields, and hopefully to open the door to future extensions.

%Figure~\ref{fig:robot} shows an example of ESGVI outperforming the standard MAP method on a difficult SLAM problem.  Our attempts to generalize ESGVI beyond Gaussian estimation naturally led us to the current paper.  We believe the tools and analysis presented herein provide a fresh way of thinking about some of the classic methods used in state estimation, and also open the door to new possibilities.

Our aim is to introduce a kind of information algebra to variational inference that not only provides a convenient and effective framework for analysis but also reveals key relationships to past work.  This algebra has its origins in the work of \citet{aitchison82} on {\em compositional data} in statistics.  Compositional data can be represented on a simplex as with probability distributions for a finite set of discrete events.  The resulting {\em Aitchison geometry} or {\em Aitchison simplex} establishes a vector space, in which vector addition is a normalized multiplication (perturbation) and scalar multiplication is a normalized exponentiation (powering).  With an appropriate inner product, the set of \acp{PDF} over a finite discrete space was formalized as a Hilbert space by \citet{pawlowsky01} and independently investigated by \citet{barfoot_phd02} and \citet{barfoot_dcss02} in their {\em stochastic algebra}.  The extension to continuous variables was first published by \citet{egozcue06} and also studied by \citet{barfoot03} for the case of finite domains.  The generalization to include probabilities and measures on the whole real line was made by \citet{vandenBoogaart2010,vandenBoogaart2014}, who introduced the term {\em Bayesian Hilbert space}.

In such a space, Bayes' venerated rule becomes
\begin{equation}\label{int:1}
   p(\mbf x|\mbf z) = p(\mbf z|\mbf x)\oplus p(\mbf x)
\end{equation}
where $\oplus$ indicates vector addition.  (The normalization inherent in the operations accounts for the marginal $p(\mbf z)$ automatically.)  Each new measurement made to refine the posterior becomes one more term added to the sum.  It is this linear feature of a Bayesian Hilbert space that makes the structure ideally suited to variational inference.

The set of Gaussians, in an appropriately extended sense, constitutes a subspace of Bayes space as do exponential families.  An arbitrary \ac{PDF} in one of these subspaces can be expressed in the simple and usual manner as a linear combination of a basis for the subspace.  The problem of variational inference can thus be expressed as the minimization of a divergence over a set of Fourier coefficients.

The linear-algebraic structure of these spaces affords us a new perspective and provides new insight.  We show, for example, that the solution to variational inference based on the \ac{KL} divergence can be viewed as an iterative projection, in the Euclidean sense, onto a given subspace.  Indeed, this algorithm is essentially a Newton-like iteration scheme to solve for the minimum of the divergence, having a form identical to the natural-gradient-descent technique of \citet{amari98}.  Moreover, using a subspace of Gaussians reproduces the recent results of \citet{barfoot_ijrr20}.

We also employ an information measure using a norm for Bayes space.  This allows for a metric to be defined on the space, which can be interpreted as the distance between two \acp{PDF}.  A (symmetric and quadratic) divergence between \acp{PDF} can be based on the distance metric.  It is notable that each step in our iterative-projection scheme is a local minimization of this divergence.

While this connection between variational inference and projection in Bayes space is exciting, there are still some open challenges around the approach.  In the current formulation, there is no guarantee that our projection scheme will result in a valid \ac{PDF}, although in practice we find the case to be so quite often, particularly with a good initial guess for the posterior estimate.  Throughout we attempt to pinpoint the specific challenges and limitations of our approach so that improvements may follow in future work.

We shall begin with an overview of Bayesian Hilbert spaces in the next section.  In \S\ref{sec:subspaces}, we discuss subspaces and bases, including exponentiated Hermite polynomials and Gaussian distributions.  The variational inference problem for the \ac{KL} divergence as viewed from the purchase of a Bayesian Hilbert space is considered in \S\ref{sec:vbi}.  The specific case of using a Gaussian subspace, that is, Gaussian variational inference, is treated in \S\ref{sec:gvi}.  Discussion is provided in \S\ref{sec:discussion} and we end with a few concluding remarks.

%%%%%%%%%%%%%%%%%%%%%%%%%%%%%%%%%%%%%%%%%%%%%%%%%%%%%%%%%%%%%%%%%%%%%%%%%%%%%%%%%%%%%%%%%%%%%%%
\section{Bayesian Hilbert Spaces}
\label{sec:bayes_space}
%%%%%%%%%%%%%%%%%%%%%%%%%%%%%%%%%%%%%%%%%%%%%%%%%%%%%%%%%%%%%%%%%%%%%%%%%%%%%%%%%%%%%%%%%%%%%%%

Let us consider some domain $\mbc{X}$ for our probability density functions \acp{PDF}, e.g., $\mathbb{R}^N$; we shall refer to $\mbf{x} \in \mbc{X}$ as the {\it state}.  A \ac{PDF} $p(\mbf{x})$ assigns a nonnegative, finite value to each element of $\mbc{X}$ such that
\begin{equation}
\label{eq:norm}
    \int_{\mbc{X}} p(\mbf{x}) \, d\mbf{x} = 1.
\end{equation}
It turns out that this condition provides challenges when it comes to defining {\em Bayes space} on an infinite domain.  As we will see, not all members of Bayes space (as we define it) will be \acp{PDF} and not all \acp{PDF} will be members of Bayes space; however, there is a large enough intersection between the two sets that Bayes space will be of practical use.  Notationally, we will use $p(\mbf{x})$ to mean a member of Bayes space throughout, indicating when the member is a valid \ac{PDF}.

We provide a lightweight explanation of Bayes space, referring to \citet{vandenBoogaart2014} for more detail.   We define the following set of functions:
\begin{equation}
    \mbc{B}^2 = \left\{ p(\mbf{x}) = c \exp( - \phi(\mbf{x}))  \, \biggl| \, 0 < c < \infty, \int_{\mbc{X}} \phi(\mbf{x})^2 \, \nu(\mbf{x}) \, d\mbf{x} < \infty \right\},
\end{equation}
where $\nu(\mbf{x})$ is an appropriate measure for $\mbc{X}$ (loosely, a weighting function); we will assume that $\nu(\mbf{x})$ is in fact a \ac{PDF} (and from $\mbc{B}^2$) throughout although this is not necessary.  Essentially, each member of $\mbc{B}^2$ is an exponentiated function from $\mbc{L}^2$, the set of square-integrable functions under our chosen measure.   Importantly, there is no requirement for $p(\mbf{x}) \in \mbc{B}^2$ to be a valid \ac{PDF}; however, if we have that $c^{-1} = \int_{\mbc{X}} \exp(-\phi(\mbf{x})) \, d\mbf{x}$, it will be so.  Moreover, not all \acp{PDF} are members of $\mbc{B}^2$ as we do not allow members to take on the value of zero anywhere in the domain\footnote{\citet{vandenBoogaart2014} explain the details around letting members of Bayes space take the value zero; our more restrictive definition sidesteps some complications.}, meaning only those \acp{PDF} that are strictly positive are contained (e.g., Gaussians and other exponential families).

We say that two members, $p_1(\mbf{x}) = c_1 \exp(-\phi_1(\mbf{x})), p_2(\mbf{x}) = c_2 \exp(-\phi_2(\mbf{x})) \in \mbc{B}^2$, are equivalent (equal) if and only if $\phi_1(\mbf{x}) = \phi_2(\mbf{x})$; in other words, the normalization constants, $c_1$ and $c_2$, need not be the same.  Under these conditions, we have that $\mbc{B}^2$ is isomorphic to $\mbc{L}^2$.

We define {\it vector addition} \citep{vandenBoogaart2010}, $\oplus:\mbc{B}^2\times\mbc{B}^2 \rightarrow \mbc{B}^2$, between two elements $p_1, p_2 \in \mbc{B}^2$ to be $p_1 \oplus p_2$:
\begin{equation}
    (\forall \mbf{x} \in \mbc{X}) \quad (p_1 \oplus p_2)(\mbf{x}) = p_1(\mbf{x})p_2(\mbf{x}) = c_1 c_2 \exp( - ( \phi_1(\mbf{x}) + \phi_2(\mbf{x}) ) ) \in \mbc{B}^2,
\end{equation}
and likewise {\it scalar multiplication}  \citep{vandenBoogaart2010}, $\cdot: \mathbb{R}\times\mbc{B}^2 \rightarrow \mbc{B}^2$, of $p \in \mbc{B}^2$ by $\alpha \in \mathbb{R}$ to be $\alpha \cdot p$:
\begin{equation}
     (\forall \mbf{x} \in \mbc{X}) \quad (\alpha \cdot p)(\mbf{x}) =(p(\mbf{x}))^\alpha = c^\alpha \exp( -\alpha \phi(\mbf{x}))  \in \mbc{B}^2.
\end{equation}
With these operations, $\mbc{B}^2$ is established as a vector space, termed a {\em Bayesian linear space}, over the field $\mathbb{R}$ \citep{vandenBoogaart2010}.  Notably, the {\em zero vector}\footnote{\citet{vandenBoogaart2014} make the point that the origin of Bayes space (i.e., the zero vector) can be shifted to be any valid member of $\mbc{B}^2$ including a \ac{PDF}, although we do not find this necessary here.} is simply any constant function, $c\exp(0)$.  Vector subtraction  \citep{vandenBoogaart2010} is defined in the usual way, $p_1 \ominus p_2 = p_1 \oplus (-1) \cdot p_2$:
\begin{equation}
     (\forall \mbf{x} \in \mbc{X}) \quad (p_1 \ominus p_2)(\mbf{x}) = \frac{c_1}{c_2} \exp( -(\phi_1(\mbf{x}) - \phi_2(\mbf{x}))) \in \mbc{B}^2.
\end{equation}
We note that subtraction, or the inverse additive operation, is equivalent to the Radon-Nikodym derivative \citep{vandenBoogaart2010}.

To turn a member of $\mbc{B}^2$ into a valid \ac{PDF} we define the {\it normalization operator}, $\norm{p}$:
\begin{equation}
    (\forall \mbf{x} \in \mbc{X}) \quad (\!\norm{p})(\mbf{x}) = \frac{p(\mbf{x})}{\int_\mbc{X} p(\mbf{z}) \, d\mbf{z}} \in \mbc{B}^2.
\end{equation}
This operation can only be applied to those members of $\mbc{B}^2$ that are equivalent to a valid \ac{PDF}; in other words, it must be that $\int_\mbc{X} p(\mbf{x}) \, d\mbf{x} < \infty$.  We will refer to the {\em subset} of $\mbc{B}^2$ whose members are equivalent to a valid \ac{PDF} as $\norm{\mbc{B}^2} \subset \mbc{B}^2$; note that this subset is not a subspace under our chosen addition and scalar multiplication operators.  As a point of order, the normalization operator is not strictly required in the establishment of $\mbc{B}^2$, only when we want to make the connection to a valid \ac{PDF}.  

As mentioned above, Bayes' rule can be rendered as $p(\mbf x|\mbf z) = p(\mbf z|\mbf x)\oplus p(\mbf x)$.  The normalizing marginal $p(\mbf z)$ is accounted for in the implied equivalence of the ``$=$'' operator.  We could also write $p(\mbf x|\mbf z) = \norm{(p(\mbf z|\mbf x)\oplus p(\mbf x))}$, which then makes the right-hand side a valid \ac{PDF} through normalization.

\paragraph{Inner Product}
We endow the vector space with an {\it inner product} \citep{vandenBoogaart2014} defined as
\begin{equation}\label{eq:ip}
    \ip{p_1}{p_2} = \frac{1}{2} \int_\mbc{X} \int_\mbc{X} \ln \left( \frac{p_1(\mbf{x})}{p_1(\mbf{y)}} \right) \ln \left( \frac{p_2(\mbf{x})}{p_2(\mbf{y})} \right) \nu(\mbf{x}) \, \nu(\mbf{y}) \, d\mbf{x} \, d\mbf{y},
\end{equation}
where $\nu(\cdot)$ is again a density function corresponding to an appropriate measure for $\mbc{X}$.  Notably, we see that because of the way the inner product is defined the normalization constants, $c_1$ and $c_2$, associated with $p_1$ and $p_2$ play no role.

Because $\nu$ is a valid \ac{PDF}, we can also write the inner product in~\eqref{eq:ip} as
\begin{equation}\label{eq:ip2}
    \ip{p_1}{p_2} = \mathbb{E}_\nu \left[ \ln p_1  \ln p_2 \right] - \mathbb{E}_\nu \left[ \ln p_1 \right] \mathbb{E}_\nu \left[ \ln p_2 \right],
\end{equation}
where $\mathbb{E}_\nu [ \cdot ]$ is the expectation with respect to $\nu$.  To be clear, when we use expectations the argument, $f(\mbf{x})$, and the measure (a \ac{PDF}), $\nu(\mbf{x})$, are defined over the same space, $\mbc{X}$:
\begin{equation}
\mathbb{E}_{\nu(\mbf{x})}[f(\mbf{x})] = \int_{\mbc{X}} f(\mbf{x}) \nu(\mbf{x}) \, d\mbf{x}, 
\end{equation}
although sometimes we will abbreviate this as $\mathbb{E}_\nu[f]$.
In this work, we shall always take the measure to be a \ac{PDF} (and from $\mbc{B}^2$); however, we shall refer to it as the measure to distinguish it from the other densities involved.  Following \citet{vandenBoogaart2014}, then, we can claim that $\mbc{B}^2$ with inner product (\ref{eq:ip}) forms a separable Hilbert space, which is referred to as a {\em Bayesian Hilbert space}.  We shall sometimes briefly refer to it as a {\em Bayesian space} or {\em Bayes space}.

\paragraph{Information and Divergence}
The {\em norm}  \citep{vandenBoogaart2014} of $p \in \mbc{B}^2$ can be taken as $\|p\| = \ip{p}{p}^{1/2}$.  Accordingly, we can define the {\em distance} between two members of $\mbc{B}^2$, $p$ and $q$, simply as $d(p,q) = \|p\ominus q\|$, which induces a metric on Bayes space.

The norm of $p$ can be used to express the information content of the PDF (if it is in $\mbc{B}^2$).  In fact, we shall define
\begin{equation}\label{eq:ip3}
    I(p) = \frac{1}{2}\|p\|^2 = \frac{1}{2}\ip{p}{p}
\end{equation}
as the {\em information}\footnote{This is different than the information of \citet{shannon48}.} in $p$.  (The reason for the factor of $\frac{1}{2}$ will become evident.)  As an example, consider $p = \mbc N(\mu,\sigma^2)$ (over the domain $\mathbb R$) and measure $\nu = \mbc N(0,1)$.  The information is $I(p) = (1+2\mu^2)/4\sigma^4$.  The smaller the variance the larger the information indicating that the PDF concentrates its probability mass more tightly about its mean; that is, we know better where to expect the state so we may say that we have more information about it.

We shall furthermore find it useful to define a {\em divergence} between two members of $\mbc{B}^2$, $p$ and $q$, as
\begin{equation}\label{eq:ip4}
    I(p\ominus q) = \frac{1}{2}\ip{p\ominus q}{p\ominus q}.
\end{equation}
This is the information contained in the difference of $p$ and $q$.  Unlike the Kullback-Leibler divergence, this divergence is symmetric in $p$ and $q$ and quadratic in Bayesian space.  Clearly, $p = q$ if and only if $I(p\ominus q) = 0$.  Geometrically, the divergence is (half) the squared Euclidean distance between $p$ and $q$ in Bayes space.

\paragraph{Stochastic Derivative}
Accompanying this algebra is a functional calculus.  Consider $p(\mbf{x}|\theta) \in \mbc{B}^2$ depending continuously on some parameter $\theta$.  We define the {\em stochastic partial derivative} of $p$ with respect to $\theta$ as \citep{barfoot03,Egozcue&al2013}
\begin{equation}\label{eq:sc}
    \frac{\eth p}{\eth \theta} = \lim_{\lambda \rightarrow 0}
    \frac{1}{\lambda}\cdot \bigl( p(\mbf{x}|\theta + \lambda) \ominus p(\mbf{x}|\theta) \bigr).
\end{equation}
Note that the result of this operation remains an element in $\mbc{B}^2$.   We can also define directional derivatives and a gradient operator  but these will not be required here.

%%%%%%%%%%%%%%%%%%%%%%%%%%%%%%%%%%%%%%%%%%%%%%%%%%%%%%%%%%%%%%%%%%%%%%%%%%%%%%%%%%%%%%%%%%%%%%%%%%
\section{Subspaces and Bases}
\label{sec:subspaces}
%%%%%%%%%%%%%%%%%%%%%%%%%%%%%%%%%%%%%%%%%%%%%%%%%%%%%%%%%%%%%%%%%%%%%%%%%%%%%%%%%%%%%%%%%%%%%%%%%%

While $\mbc{B}^2$ is an infinite-dimensional space, it contains infinitely many finite-dimensional subspaces.  We can in fact build a subspace $\mbc{Q}$ by taking the span of a set of $M$ vectors $B = \{b_1,\ldots,b_M\}$, namely,
\begin{equation}
    \mbc{Q} = \mbox{span}\left\{ b_1, \ldots, b_M \right\}.
\end{equation}
If we choose $B$ to be linearly independent, it will form a basis for $\mbc{Q}$.  We can accordingly write every vector $q$ in $\mbc{Q}$ as a linear combination of $B$, i.e.,
\begin{equation}
    q = \bigoplus_{m=1}^M \alpha_m\cdot b_m,
\end{equation}
where $\alpha_m \in \mathbb{R}$ are unique.  We use the notation $\bigoplus_{m=1}^M \alpha_m \cdot b_m$ to mean $\alpha_1 \cdot b_1 \oplus \cdots \oplus \alpha_M \cdot b_M$, paralleling $\sum_{m=1}^M$ for normal addition.

As a shorthand, we will denote $\mbf{b} = \bbm b_1 & b_2 & \cdots & b_M\ebm^T$ as the basis.  The inner products between all pairs of basis vectors form the {\em Gram matrix},
\begin{equation}
    \ip{\mbf{b}}{\mbf{b}} = \bbm \ip{b_m}{b_n}\ebm,
\end{equation}
where $(m,n)$ are the indices of the matrix entries.  We furthermore have an orthonormal basis if $\ip{b_m}{b_n} = \delta_{mn}$, the Kronecker delta, in which case $\ip{\mbf b}{\mbf b} = \mbf 1$, the identity matrix.

\begin{figure}[t]
\centering
\includegraphics[width=0.65\textwidth]{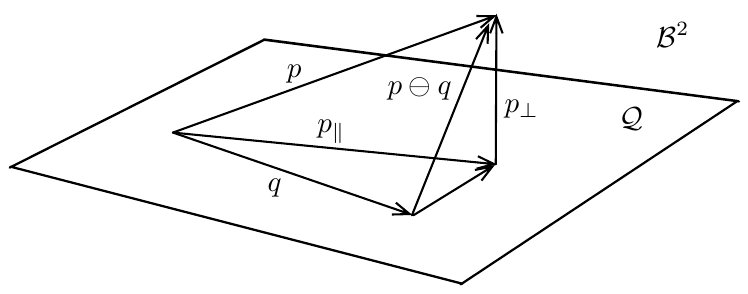}
\caption{Projection onto a subspace, $\mathcal{Q}$, of the Bayesian Hilbert space, $\mbc{B}^2$}
\label{fig:proj}
\end{figure}

\subsection{Projections}
\label{sec:projections}

Given a subspace $\mbc Q$ of $\mbc{B}^2$ and $p \in \mbc{B}^2$, the $q^\star \in \mbc Q$ that minimizes the distance to, as well as the divergence~\eqref{eq:ip4} from, $p$ is the projection of $p$ onto $\mbc Q$, that is,
\begin{equation}\label{proj:1}
    q^\star = \underset{\mbc Q}{\text{proj}}\, p.
\end{equation}
As in Euclidean geometry, we can view $p$ as being decomposed into a component $p_\|$ lying in $\mbc Q$ and a component $p_\perp$ perpendicular to it; therefore $q^\star = p_\|$ (see Figure~\ref{fig:proj}).

The coordinates of $q^\star$ can be calculated as
\begin{equation}\label{proj:2}
    \mbs\alpha^\star = \ip{\mbf b}{\mbf b}^{-1}\ip{\mbf b}{p}.
\end{equation}
We may also write the projection as an {\it outer-product} operation on $p$, as detailed in Appendix~\ref{sec:op}.

\begin{figure}[t]
\centering
\includegraphics[width=0.8\textwidth]{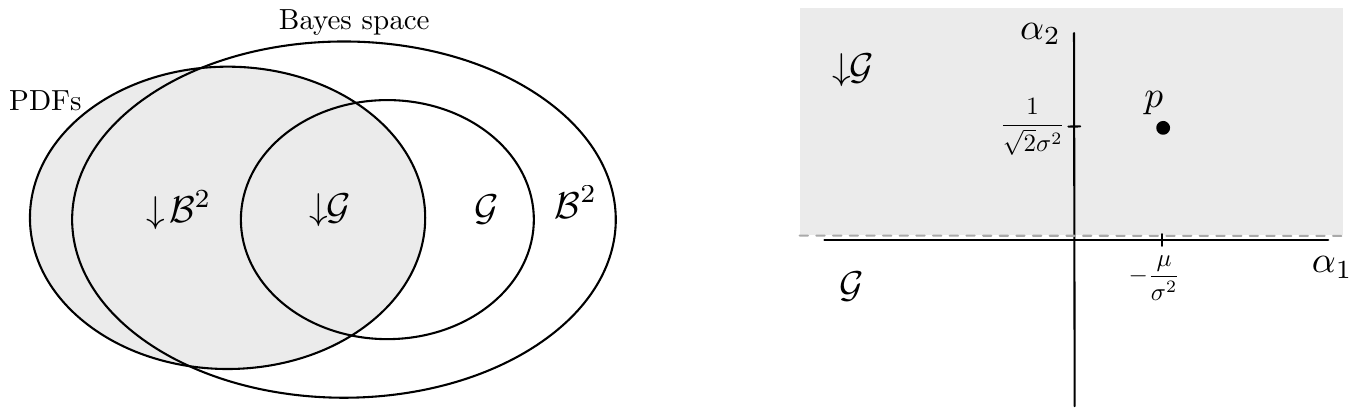}
\caption{On the left is a depiction of the relationships between Bayes space, $\mbc{B}^2$, and the set of all \acp{PDF}.  We see the subset of strictly positive \acp{PDF}, $\norm{\mbc{B}^2}$, the indefinite-Gaussian subspace, $\mbc{G}$, and the positive-definite-Gaussian subset, $\norm{\mbc{G}}$.  The right image shows how a valid Gaussian \ac{PDF} can be viewed as a point in a plane with coordinates that depend on its mean $\mu$ and variance $\sigma^2$; only the open upper-half plane admits valid Gaussian \acp{PDF} since we must have $\sigma^2 > 0$}
\label{fig:subsets2}
\end{figure}

\subsection{Example:  One-Dimensional Gaussian}
To make the concept of Bayes space more tangible, consider the canonical one-dimensional Gaussian \ac{PDF} defined over $x \in \mathbb{R}$:
\begin{equation}
\label{eq:gauss1d}
p(x) = \frac{1}{\sqrt{2 \pi \sigma^2}} \exp\left( -\frac{1}{2} \frac{(x-\mu)^2}{\sigma^2} \right),
\end{equation}
where $\mu$ is the mean and $\sigma^2$ the variance.  In the language of $\mbc{B}^2$, we can write this as
\begin{equation}
p(x) = \underbrace{\left( -\frac{\mu}{\sigma^2}\right)}_{\alpha_1} \cdot \underbrace{\exp(-x)}_{b_1} \oplus \underbrace{\left(\frac{1}{\sqrt{2}\sigma^2}\right)}_{\alpha_2} \cdot \underbrace{\exp\left( -\frac{(x^2 - 1)}{\sqrt{2}} \right)}_{b_2} = \alpha_1 \cdot b_1 \oplus \alpha_2 \cdot b_2.
\end{equation}
In other words, every Gaussian can be written as a linear combination of the two vectors, $b_1$ and $b_2$, where the coefficients, $\alpha_1$ and $\alpha_2$, depend on the mean and variance.  Note, we can neglect the normalizing constant as equivalence is implied in the ``$=$'' operator.

The choice of $b_1$ and $b_2$ is not arbitrary in this example.  They constitute the first two basis vectors in an orthonormal basis for $\mbc{B}^2$, which can be established using the {\em probabilist's Hermite polynomials}; Appendix~\ref{sec:herm_r} provides the details of this {\em Hermite basis}. In fact, we can define a new space $\mbc{G}$ as the span of these two basis vectors:
\begin{equation}
\mbc{G} = \mbox{span}\left\{ b_1, b_2 \right\},
\end{equation}
which is a subspace of $\mbc{B}^2$.  Importantly, every Gaussian \ac{PDF} of the form in~\eqref{eq:gauss1d} is a member of $\mbc{G}$, but not every member of $\mbc{G}$ is a valid Gaussian \ac{PDF}.  Only those members of $\mbc{G}$ that have $\sigma^2 > 0$ are valid Gaussian \acp{PDF}.  We shall refer to $\mbc{G}$ as the {\em indefinite-Gaussian subspace of $\mbc{B}^2$} while $\norm{\mbc{G}} \subset \mbc{G}$ will denote the {\em positive-definite-Gaussian subset}.  Figure~\ref{fig:subsets2} shows the relationships between the various spaces and how we can view a Gaussian as a point in $\mbc{G}$.

\subsection{Example:  Projecting to a Gaussian}
\label{subsec:num_ex}

Let us consider a simple one-dimensional, nonlinear estimation problem as a numerical example motivated by the type of inverse-distance nonlinearity found in a stereo camera model.  This same experiment (with the same parameter settings) was used as a running example by \citet[\S4]{barfoot17}.  We assume that our true state is drawn from a Gaussian prior:
\begin{equation}
    x  \sim \mathcal{N}(\mu_p, \sigma_p^2).
\end{equation}
We then generate a measurement according to
\begin{equation}
    z = \frac{fb}{x} + n, \quad n \sim \mathcal{N}(0,\sigma_r^2),
\end{equation}
where $n$ is measurement noise.  The numerical values of the parameters used were
\begin{equation}
\begin{gathered}
\label{eq:stereoexpparam}
    \mu_p = 20 \mbox{ [m]}, \quad \sigma_p^2 = 9 \mbox{ [m$^2$]}, \\
    f = 400 \mbox{ [pixel]}, \quad b = 0.1 \mbox{ [m]}, \quad \sigma_r^2 = 0.09 \mbox{ [pixel$^2$]}. 
\end{gathered}
\end{equation}
The true posterior is given by
\begin{equation}
    p(x|z) = \norm{\exp(-\phi(x))}, \quad \phi(x) = \underbrace{\frac{1}{2} \frac{(x -
        \mu_p)^2}{\sigma_p^2}}_{\mbox{prior}} + \underbrace{\frac{1}{2} \frac{\left(z - \frac{fb}{x}\right)^2}{\sigma_r^2}}_{\mbox{measurement}}.
\end{equation}
This problem can also be viewed as the correction step of the Bayes filter \citep{jazwinsky70}:  Start from a prior and correct it based on the latest (nonlinear) measurement.

\begin{figure}[t]
\centering
\includegraphics[width=0.7\textwidth]{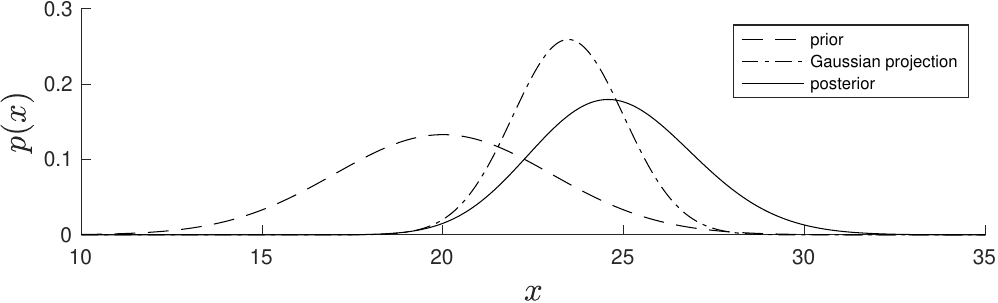}\\ \medskip
\includegraphics[width=0.715\textwidth]{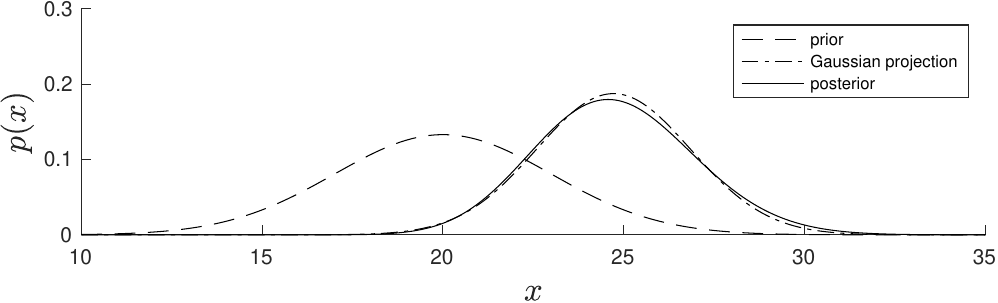}\hspace*{0.00in}
\caption{An example of projecting a non-Gaussian posterior onto the indefinite-Gaussian subspace.  The top panel shows the case where the measure associated with $\mbc{B}^2$ was chosen to be the same as the (Gaussian) prior, $\nu(x) = \mathcal{N}(20,9)$.  The bottom panel does the same with a Gaussian measure selected to be closer to the posterior, $\nu(x) = \mathcal{N}(24,4)$.  We see that the Gaussian projection of the posterior is much closer to the true posterior in the bottom case}
\label{fig:exp1}
\end{figure}

We seek to find $q(x) \in \norm{\mbc{G}}$ that is a good approximation to the true posterior $p(x|z)$.  To do this we will simply project the posterior onto the indefinite-Gaussian subspace, using the method described in Section~\ref{sec:projections}, and then normalize the result.  Figure~\ref{fig:exp1} shows the results of doing this for two cases that differ only in the measure $\nu$ that we associate with Bayes space.   The expectations used in the projections were computed with generic numerical integration although, as discussed by \citet{barfoot_ijrr20}, there are several other options including Gaussian quadrature.  In the top case, the measure is chosen as the prior estimate, while in the bottom case it is chosen to be closer to the posterior.  In both cases, we see that our projection method produces a valid \ac{PDF}, but in the bottom case the result is much closer to the true posterior.  This simple example provides motivation for the main point of this paper, which is that to use the tools of Bayes space effectively, we will seek to iteratively update the measure used to carry out our projections such that we can best approximate a posterior.  Intuitively, this makes sense since the measure is providing a weighting to different parts of $\mathbb{R}$ so we would like to choose it to pay close attention where the posterior ends up.

%%%%%%%%%%%%%%%%%%%%%%%%%%%%%%%%%%%%%%%%%%%%%%%%%%%%%%%%%%%%%%%%%%%%%%%%%%%%%%%%%%%%%%%%%%%%%%%
\section{Variational Bayesian Inference}
\label{sec:vbi}
%%%%%%%%%%%%%%%%%%%%%%%%%%%%%%%%%%%%%%%%%%%%%%%%%%%%%%%%%%%%%%%%%%%%%%%%%%%%%%%%%%%%%%%%%%%%%%%

Motivated by the example in Section~\ref{subsec:num_ex}, we shall now address the problem of variational Bayesian inference using the algebraic tools of Bayes space.  
\subsection{Variation on the Kullback-Leibler Divergence}

In variational Bayesian inference, we seek to find an approximation, $q$, from some family of distributions constituting a subspace $\mbc{Q}$, to the true Bayesian posterior $p \in \mbc{B}^2$.  In general,
\begin{equation}
    \mbc{Q} \subseteq \mbc{B}^2,
\end{equation}
where equality will always ensure that $q=p$ will match the posterior exactly.  But $\mbc{B}^2$ is infinite-dimensional and, in practice, $\mbc{Q} \subset \mbc{B}^2$ is a finite-dimensional subspace.

There are many possible {\em divergences} that can be defined to characterize the `closeness' of $q$ to $p$ including the Kullback-Leibler (\ac{KL}) divergence \citep{kullback51}, Bregman divergence \citep{Bregman1967}, Wasserstein divergence/Earth mover's distance \citep{monge1781} and R\'enyi divergence \citep{Renyi1961}.  We shall focus on the \ac{KL} divergence, which is defined as
\begin{equation}
    \mbox{KL}(q || p) = - \int_\mbc{X} q(\mbf{x}) \ln \left( \frac{p(\mbf{x} | \mbf{z})}{q(\mbf{x})} \right) \, d\mbf{x} = -\mathbb{E}_q [ \ln p - \ln q].
\end{equation}
Sometimes the reverse of this is used:  $\mbox{KL}(p || q)$.  Note, we show the divergence with respect to the posterior, $p(\mbf{x} | \mbf{z})$, but in practice during the calculations we use that $p(\mbf{x} | \mbf{z}) = p(\mbf{x},\mbf{z}) / p(\mbf{z}) = \norm{p(\mbf{x},\mbf{z})}$ since the joint likelihood is easy to construct and then the $p(\mbf{z})$ can be dropped for it does not result in a KL term that depends on $\mbf{x}$.  We will generically use $p$ in what follows to keep the notation clean.

\subsection{KL Gradient}
We assume a basis $B = \{b_1,b_2\cdots b_M\}$ for $\mbc Q$ and we write $q$ as
\begin{equation}
    q = \norm{} \bigoplus_{m=1}^M \alpha_m\cdot b_m.
\end{equation}
We desire to minimize the \ac{KL} divergence with respect to the coordinates $\alpha_m$.  The gradient of $\mbox{KL}(q || p)$ can be computed as follows:
\begin{equation}\label{kl:1}
    \frac{\partial\mbox{KL}}{\partial\alpha_n}
        = -\int_{\mbc X} \left(\frac{\partial q}{\partial\alpha_n}(\ln p - \ln q) - q\frac{\partial \ln q}{\partial\alpha_n}\right)d\mbf x.
\end{equation}
Exploiting (\ref{fish:3}) and (\ref{fish:4}), this reduces to
\begin{equation}
    \frac{\partial\mbox{KL}}{\partial\alpha_n}
        = -\mathbb E_q[\ln b_n(\ln p - \ln q)] + \mathbb E_q[\ln b_n]
            \mathbb E_q[\ln p - \ln q]
        = -\ip{b_n}{p\ominus q}_q
\end{equation}
or, collecting these in matrix form,
\begin{equation}\label{kl:2}
    \frac{\partial\mbox{KL}}{\partial\mbs\alpha^T} = -\ip{\mbf b}{p\ominus q}_q,
\end{equation}
where $\mbf{b} = \bbm b_1 & b_2 & \cdots & b_M \ebm^T$.  Implicit in this statement is that when employed as the measure, we have normalized the current approximation, $\norm{q^{(i)}}$, since we always take the measure to be a valid \ac{PDF}.
The necessary condition for a minimum of the \ac{KL} divergence is that the gradient is zero.  Newton's method suggests the manner in which we might iteratively solve for the optimal distribution.  Following the established procedure, the iteration for the coordinates is given by
\begin{equation}\label{kl:3}
    \mbs\alpha^{(i+1)} = \mbs\alpha^{(i)} + {\mbf H^{(i)}}^{-1}\ip{\mbf b}{p\ominus q^{(i)}}_{q^{(i)}},
\end{equation}
where $\mbf H$ is the Hessian of the \ac{KL} divergence.

\subsection{KL Hessian}
The $(m,n)$ entry of the Hessian is
\begin{equation}
    \frac{\partial^2\mbox{KL}}{\partial\alpha_m\partial\alpha_n}
        = -\frac{\partial}{\partial\alpha_m}\ip{b_n}{p\ominus q}_q.
\end{equation}
This differentiation must take into account the effect of the `measure' $q$.  The product rule applies here and we can break down the differentiation as
\begin{equation}\label{kl:4}
    \frac{\partial^2\mbox{KL}}{\partial\alpha_m\partial\alpha_n}
        = -\left(\frac{\partial}{\partial\alpha_m}\ip{b_n}{p\ominus q}\right)_q
            - \ip{b_n}{p\ominus q}_{\partial q/\partial\alpha_m},
\end{equation}
the first term of which is to be read as the derivative of the inner product holding the measure fixed and the second of which deals with the derivative of the measure while holding the arguments of inner product fixed.  The first term is
\begin{equation}\label{kl:5}
    \left(\frac{\partial}{\partial\alpha_m}\ip{b_n}{p\ominus q}\right)_q
        = \left(\frac{\partial}{\partial\alpha_m}\ip{b_n}{p}
        - \frac{\partial}{\partial\alpha_m}\sum_k\alpha_k\ip{b_n}{b_k}\right)_q
        = -\ip{b_n}{b_m}_q = -\ip{b_m}{b_n}_q.
\end{equation}
As shown in Appendix~\ref{sec:ipderiv}, the second becomes
\begin{equation}\label{kl:6}
    \ip{b_n}{p\ominus q}_{\partial q/\partial\alpha_m} = \ip{b_n}{\frac{\partial\ln q}{\partial\alpha_m}\cdot(p\ominus q)}_q - \mathbb E_q[\ln p - \ln q]\ip{b_m}{b_n}_q.
\end{equation}
We advise that the coefficient $\partial\ln q/\partial\alpha_m$ of $p\ominus q$ is in fact a function of the state and as such cannot be transferred to the other argument of the inner product as would be possible for a scalar in the field $\mathbb R$.  We also recognize the factor of the last term as $\mbox{KL}(q||p)$.  Therefore, substituting (\ref{kl:5}) and (\ref{kl:6}) into (\ref{kl:4}) yields
\begin{equation}\label{kl:7}
    \frac{\partial^2\mbox{KL}}{\partial\alpha_m\partial\alpha_n}
        = \left(1 - \mbox{KL}(q||p)\right)\ip{b_m}{b_n}_q
        - \ip{b_n}{\frac{\partial\ln q}{\partial\alpha_m}\cdot(p\ominus q)}_q.
\end{equation}
We observe that the second term on the right-hand side is symmetric in the indices as the substitution of (\ref{fish:4}) will attest.  In matrix form, the Hessian is
\begin{equation}\label{kl:8}
    \mbf H = \frac{\partial^2\mbox{KL}}{\partial\mbs\alpha^T\partial\mbs\alpha}
        = \left(1 - \mbox{KL}(q||p)\right)\mbf I_{\mbs\alpha}
        - \ip{\mbf b}{\frac{\partial\ln q}{\partial\mbs\alpha^T}\cdot(p\ominus q)}_q,
\end{equation}
where $\mbf I_{\mbs\alpha}$ is the {\em Fisher information matrix} (FIM) or {\em Gram matrix} and is described in detail in Appendix~\ref{sec:gram}.
Newton's method (\ref{kl:3}) can now be implemented.  But the Hessian bears a closer look.

The Hessian can also be explicitly written as
\begin{multline}
    \frac{\partial^2\mbox{KL}}{\partial\alpha_m\partial\alpha_n} = \ip{b_m}{b_n}_q
        - \mathbb E_q[\ln b_m\ln b_n(\ln p - \ln q)] \\ + \mathbb E_q[\ln b_m\ln b_n]\mathbb E_q[\ln p - \ln q] 
        + \mathbb E_q[\ln b_n]\mathbb E_q[\ln b_m(\ln p - \ln q)] \\ + \mathbb E_q[\ln b_m]\mathbb E_q[\ln b_n(\ln p - \ln q)] 
        - 2\mathbb E_q[\ln b_m]\mathbb E_q[\ln b_n]\mathbb E_q[\ln p - \ln q],
\end{multline}
the terms of which can be collected as
\begin{equation}\label{kl:8a}
    \frac{\partial^2\mbox{KL}}{\partial\alpha_m\partial\alpha_n} = \ip{b_m}{b_n}_q
        + \ip{-b_{mn} + \mathbb E_q[\ln b_n]\cdot b_m + \mathbb E_q[\ln b_m]\cdot b_n}{p\ominus q}_q,
\end{equation}
where $b_{mn} = \exp(\ln b_m\ln b_n)$.  The symmetry in the Hessian is plainly evident in this version.

\subsection{Iterative Projection}
\begin{figure}[t]
\centering
\ifthenelse{\boolean{narrow}}
{
\includegraphics[width=0.6\textwidth]{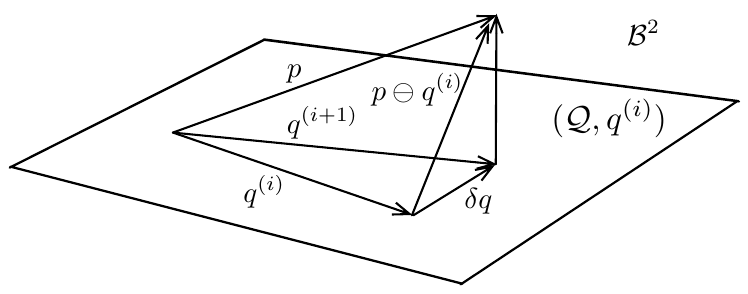}
}
{
\vspace*{-0.1in}
\includegraphics[width=0.6\textwidth]{iterproj2.pdf}
\vspace*{-0.1in}
}
\caption{Iterative projection onto a sequence of Bayesian Hilbert spaces, $(\mathcal{Q},q^{(i)})$}
\label{fig:iterproj}
\end{figure}

In the vicinity of the optimal distribution, with a sufficiently large subspace $\mbc Q$, we may expect $p\ominus q$ to be small almost everywhere.  This makes all the terms in the Hessian of first order except $\mbf I_{\mbs\alpha}$, which is of zeroth order.  The gradient (68) is also of first order.
%$p\ominus q$ is small and in fact we may express this as $p\ominus q = O(\varepsilon)$ in the sense that $\ln p - \ln q = O(\varepsilon)$ almost everywhere.  Likewise $\mbox{KL}(q||p) = O(\varepsilon)$.  The $\mbox{KL}$ term as well as the last term in (\ref{kl:8}) and, equivalently, the second term in (\ref{kl:8a}) are accordingly of $O(\varepsilon)$. The gradient (\ref{kl:2}) is of the same order.  
Thus to keep Newton's descent to this order, we may approximate the Hessian as $\mbf H \simeq \mbf I_{\mbs\alpha}$ and the iterative procedure (\ref{kl:3}) becomes simply
\begin{equation}\label{kl:9}
    \mbs\alpha^{(i+1)} = \mbs\alpha^{(i)} + {\mbf I_{\mbs\alpha}^{(i)}}^{-1}\ip{\mbf b}{p\ominus q^{(i)}}_{q^{(i)}}.
\end{equation}
However, as $q^{(i)} = \norm \bigoplus_m \alpha_m^{(i)}\cdot b_m$,
\begin{equation}
    \ip{\mbf b}{p\ominus q^{(i)}}_{q^{(i)}} = \ip{\mbf b}{p}_{q^{(i)}}
        - \ip{\mbf b}{\mbf b}_{q^{(i)}}\mbs\alpha^{(i)}
        = \ip{\mbf b}{p}_{q^{(i)}} - \mbf I_{\mbs\alpha}^{(i)}\mbs\alpha^{(i)}.
\end{equation}
Hence (\ref{kl:9}) becomes
\begin{equation}\label{kl:10}
    \mbs\alpha^{(i+1)} = {\mbf I_{\mbs\alpha}^{(i)}}^{-1}\ip{\mbf b}{p}_{q^{(i)}}.
\end{equation}
The iterative update to $q$ is $q^{(i+1)} = \norm \bigoplus_m \alpha_m^{(i+1)}\cdot b_m$.  %We may then express (\ref{kl:10}) using the application of the kernel of $\mbc Q$,
%\begin{equation}\label{kl:11}
%    q^{(i+1)} =
%        \left(\mbf b\rangle {\mbf I_{\mbs\alpha}^{(i)}}^{-1} \langle \mbf b\right)\circledast
 %       p,
%\end{equation}
%which describes a projection onto $\mbc Q$ using measure $q^{(i)}$.  
That is, the procedure can be viewed as an {\em iterative projection},
\begin{equation}\label{kl:12}
    q^{(i+1)} = \underset{(\mbc Q, q^{(i)})}{\norm{} \text{proj}} p,
\end{equation}
where we explicitly indicate that we normalize the result as the output of our algorithm should be a PDF.
Figure~\ref{fig:iterproj} depicts the scheme.  The procedure is essentially the application of Newton's method on the \ac{KL} divergence with the Hessian approximated as the \ac{FIM}.  This is precisely the approximation made in natural gradient descent \citep{amari98}.  In our Bayesian space, the operating point of the Newton step becomes the measure for the inner product.  This highlights a key aspect of using the algebra associated with a Bayesian space.  It recognizes the dual role of $q$: On the one hand it is the approximating PDF and on the other it serves as a measure that weights the difference between the approximation and the approximated.

Convergence of iterative projection is guaranteed if the Hessian is positive-definite.  Provided that the subspace is large enough, we can expect convergence when we begin in a neighborhood of  optimal $q$ where the first-order terms in the Hessian are sufficiently small.

It is notable that each step of the iterative projection is equivalent to the local minimization of the divergence $I(p\ominus q)$ with the measure fixed at $q^{(i)}$ because
\begin{equation}
    I\left(p\ominus( q^{(i)}\oplus\delta q)\right) = I(p\ominus q^{(i)})
        + \delta\mbs\alpha^T\left(\frac{\partial I}{\partial\mbs\alpha^T}\right)_{q^{(i)}}
        + \frac{1}{2}\delta\mbs\alpha^T
            \left(\frac{\partial^2 I}{\partial\mbs\alpha^T\partial\mbs\alpha}\right)_{q^{(i)}}
            \delta\mbs\alpha,
\end{equation}
where $\delta\mbs\alpha = \mbs\alpha^{(i+1)} - \mbs\alpha^{(i)}$ and
\begin{equation}
    \left(\frac{\partial I}{\partial\mbs\alpha^T}\right)_{q^{(i)}}
        = -\ip{\mbf b}{p\ominus q^{(i)}}_{q^{(i)}}, \quad
    \left(\frac{\partial^2 I}{\partial\mbs\alpha^T\partial\mbs\alpha}\right)_{q^{(i)}}
        = \ip{\mbf b}{\mbf b}_{q^{(i)}} \equiv \mbf I_{\mbs\alpha}^{(i)},
\end{equation}
which are identical to the linearized forms for the \ac{KL} divergence.

Throughout this section we have assumed that the basis $B$ remains constant across iterations, but this need not be the case.  We may also choose to update the basis along with the measure to maintain, for example, orthonormality.  This is explored in the next example and further in Section~\ref{sec:gvi} on Gaussian variational inference.

\begin{figure}[p]
\centering
\includegraphics[width=0.7\textwidth]{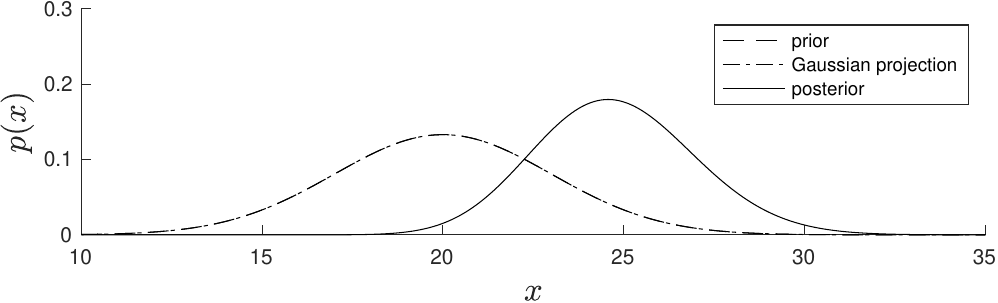}
\includegraphics[width=0.7\textwidth]{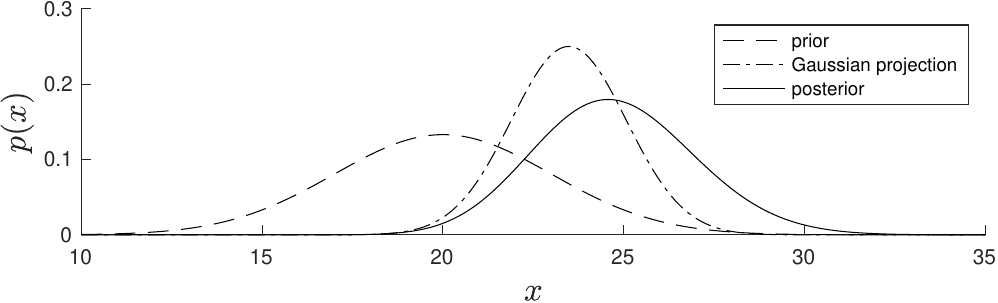}
\includegraphics[width=0.7\textwidth]{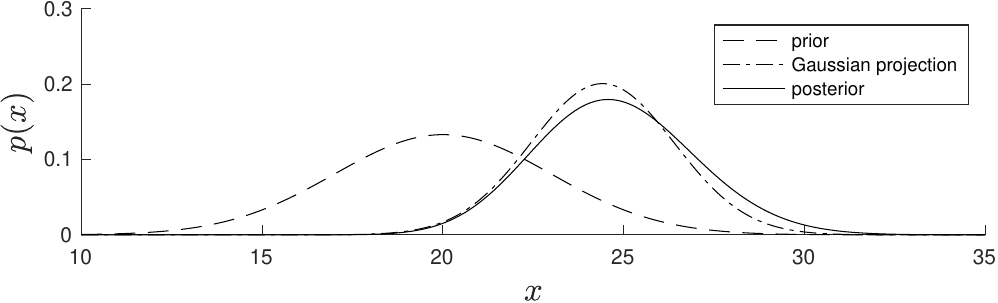}
\includegraphics[width=0.7\textwidth]{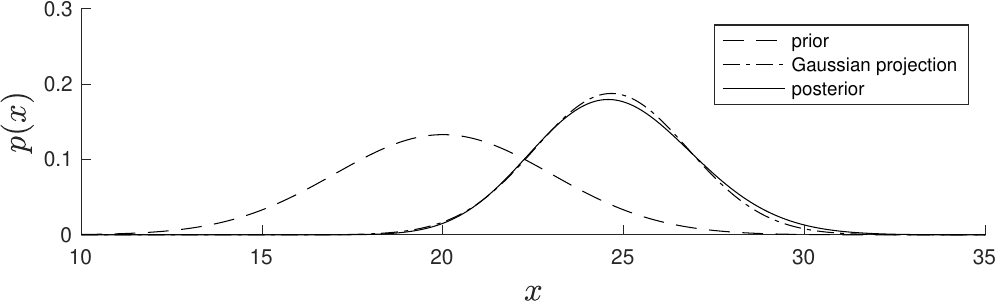}
\includegraphics[width=0.7\textwidth]{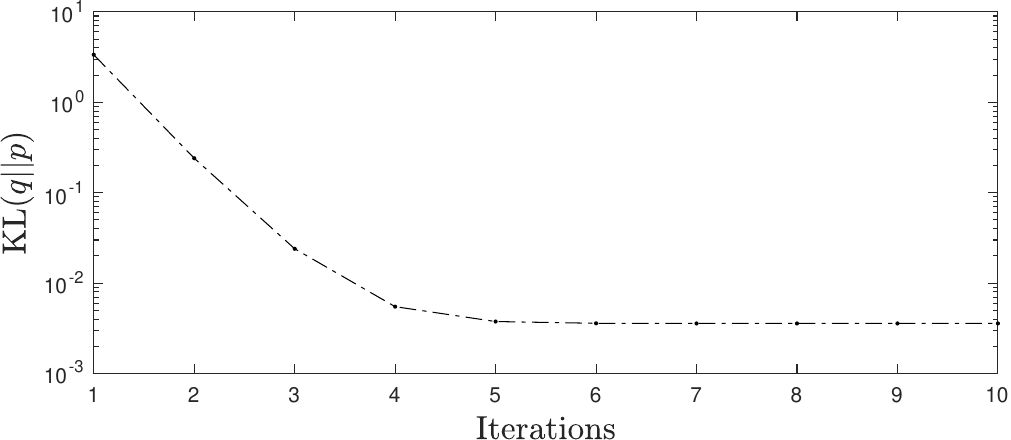}
\caption{Example of iterative projection onto the indefinite-Gaussian subspace spanned by two Hermite basis functions, where the measure is taken to be the estimate $q^{(i)}$ at the previous iteration and the basis reorthogonalized at each iteration as described in \S\ref{sec:gvi}.  The estimate was initialized to the prior (first panel) and then iteratively updated (next three panels).  The last panel shows the KL divergence between the estimate and the true posterior for 10 iterations, with convergence occurring at approximately 5 iterations}
\label{fig:exp2}
\end{figure}

\subsection{Example:  Iteratively Projecting to a Gaussian}

\label{sec:num_ex}

In the example of Section~\ref{subsec:num_ex}, we saw that selecting a measure that was closer to the posterior resulted in a projection that was also closer to the posterior.  We now redo this example using the iterative projection concepts from this section.  We will still project onto the indefinite-Gaussian subspace and employ a Gaussian measure, only now with each iteration the measure will be taken to be the (normalized) projection from the previous iteration.

We initialized the estimate to the prior, which corresponds to the first panel in Figure~\ref{fig:exp2}.  The next three panels show subsequent iterations of the estimate.  The last panel shows the KL divergence between the estimate and the true posterior for 10 iterations.  We see the estimate converged in a few iterations.

Note that as the measure changes from one iteration to the next, we then have to update the basis to retain the desired orthogonality.  This can be accomplished by using the reparameterization `trick'  (see Appendix~\ref{sec:herm_r}) to adjust the basis to be orthogonal with respect to the current Gaussian measure.

\subsection{Exploiting Sparsity}
\label{sec:sparsity}

One of the major advantages of thinking of $\mbc{B}^2$ as a vector space with the definition of vector addition $\oplus$ is that Bayesian inference in general can be viewed as the addition of vectors.  Consider the posterior $p(\mbf{x} | \mbf{z})$ where $\mbf{z}$ are some measurements.  Bayes' rule states that
\begin{equation}
    p(\mbf{x} | \mbf{z}) = \frac{p(\mbf{z} | \mbf{x}) p(\mbf{x})}{p(\mbf{z})} = p(\mbf{z} | \mbf{x}) \oplus p(\mbf{x}),
\end{equation}
where $p(\mbf{x})$ is a prior, $p(\mbf{z} | \mbf{x})$ is a measurement factor and, as mentioned earlier, we needn't introduce the normalization constant $p(\mbf{z})$ explicitly when writing the posterior as a vector addition in Bayesian space.  To be clear, addition is defined for two members of the same Bayes space and here we interpret $p(\mbf{z} | \mbf{x})$ as a function of $\mbf{x}$ since $\mbf{z}$ is a constant (e.g., a known measurement).

If we have several measurements that are statistically independent, then this can be factored as
\begin{equation}
    p(\mbf{x} | \mbf{z}) = p(\mbf{x}) \oplus \bigoplus_{k=1}^K p(\mbf{z}_k | \mbf{x}_k),
\end{equation}
where $\mbf{x}_k = \mbf{P}_k \mbf{x}$ is a subset of the variables in $\mbf{x}$, $\mbf{P}_k$ is a projection matrix, and $\mbf{z}_k$ is the $k$th measurement.  This expresses sparsity in the state description and in the measurements.  To keep the notation economical, we shall simply write
\begin{equation}\label{spar:1}
    p = \bigoplus_{k=0}^K p_k,
\end{equation}
where $p$ is the posterior and the $p_k$ comprise the prior and the measurements, corresponding to statistically independent data.  In other words, the factorization becomes a summation in the Bayesian space $\mbc{B}^2$.

Now consider our projective approach to inference.  As usual, given a subspace $\mbc Q \subset \mbc{B}^2$, the optimal estimate to (\ref{spar:1}) is given by
\begin{equation}
    q^\star = \underset{\mbc{Q}}{\mbox{\normalfont proj}} \, p
    = \underset{\mbc{Q}}{\mbox{\normalfont proj}} \,\bigoplus_{k=0}^K p_k
    = \bigoplus_{k=0}^K \underset{\mbc{Q}}{\mbox{\normalfont proj}} \, p_k.
\end{equation}
That is, the projection of the sum is the sum of the projections.  Each individual projection can be done separately because we are in a linear space.  This is of enormous practical advantage because it means that we do not need all of $\mbc{Q}$ to represent each projection.

We can see this more clearly by defining $\mbc{B}^2_k \subset \mbc{B}^2$ as the subspace corresponding to the variables $\mbf x_k$.  Then
\begin{equation}
    p = \bigoplus_{k=0}^K p_k \in \mbc{B}^2_0\oplus\mbc{B}^2_1\oplus \cdots \oplus\mbc{B}^2_K
        \subseteq \mbc{B}^2.
\end{equation}
In other words, $p$ is contained in the direct sum of the subspaces $\mbc{B}^2_k$.  Each constituent part $p_k$ may be confined to a smaller subspace of $\mbc{B}^2$, depending on the variable dependencies in each term.

If we wish to project $p_k \in \mbc{B}^2_k$ onto $\mbc Q$ it will suffice to consider the projection on just $\mbc Q_k = \mbc{B}^2_k \cap \mbc Q$, i.e.,
\begin{equation}
    \underset{\mbc{Q}}{\mbox{\normalfont proj}} \, p_k = \underset{\mbc{Q}_k}{\mbox{\normalfont proj}} \, p_k.
\end{equation}
The subspace $\mbc Q_k$ may, and ideally would, be smaller than $\mbc Q$.  We may refer to $\mbc{Q}_k$ as the {\em marginal subspace} of $\mbc{Q}$ with respect to the subset of variables $\mbf{x}_k$.

\begin{figure}[t]
\centering
\includegraphics[width=0.8\textwidth]{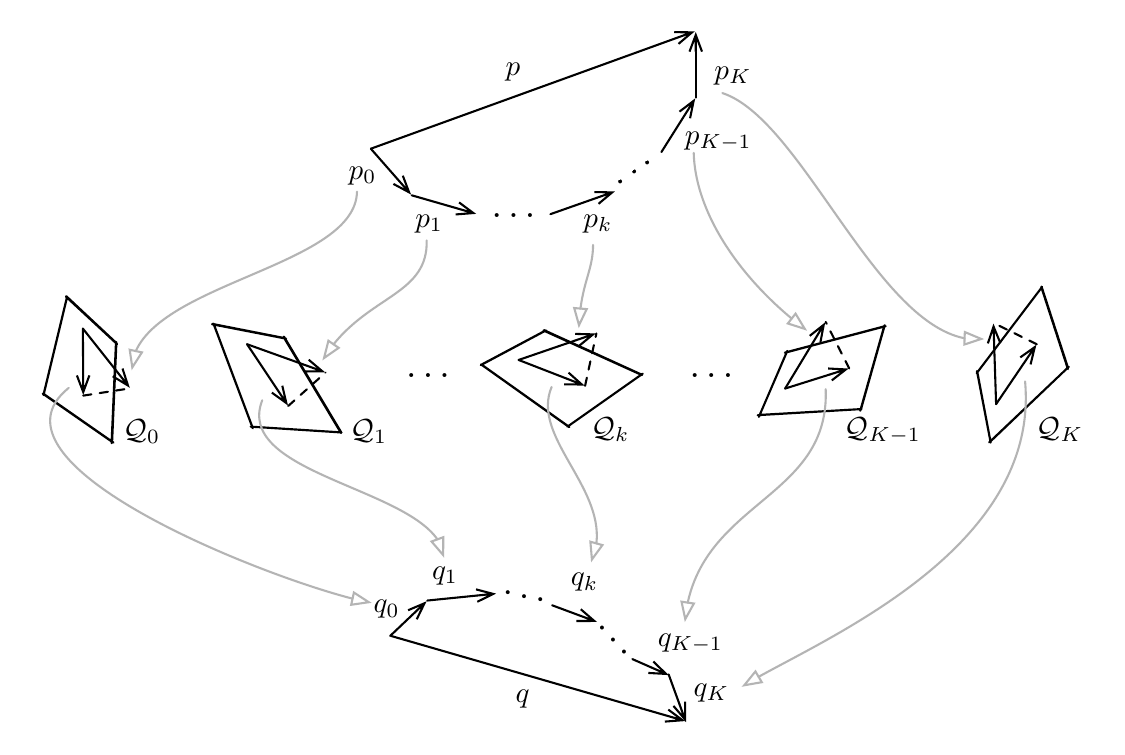}
\caption{Exploiting sparsity by projecting individual measurements onto marginal subspaces, $\mathcal{Q}_k$, and then recombining the results}
\label{fig:sparsity}
\end{figure}

Therefore, the optimal estimate will be given by
\begin{equation}
    q^\star = \underset{\mbc{Q}}{\mbox{\normalfont proj}} \, p
        = \bigoplus_{k=0}^K \underset{\mbc{Q}_k}{\mbox{\normalfont proj}} \, p_k.
\end{equation}
This means that we can project the \ac{PDF} associated with each measurement onto a smaller subspace and simply add up the estimates, lifting the overall estimate up into a potentially much larger space.  Naturally, when employed in practice we will normalize $q^\star$ to ensure our algorithm outputs a valid PDF.  The decomposition and reconstitution is illustrated in Figure~\ref{fig:sparsity}.  Just as with the total posterior, we may describe $q^\star$ as being an element of a direct sum of the individual subspaces of $\mbc Q$, i.e.,
\begin{equation}
    q^\star \in \mbc{Q}_0 \oplus \mbc{Q}_1 \oplus \cdots \oplus \mbc{Q}_K \subseteq \mbc{Q}.
\end{equation}
The subspace sum may be substantially smaller than $\mbc Q$ but again it will depend on the variable dependencies of each term.

This is the key result that allows most practical inference frameworks to function in a tractable way.  Depending on the chosen basis for $\mbc Q$, many of the coordinates can potentially be zero and thus it will not be necessary to waste effort computing them or space storing them.

%%%%%%%%%%%%%%%%%%%%%%%%%%%%%%%%%%%%%%%%%%%%%
\section{Application: Iterative Projection for Multivariate Gaussians}
\label{sec:gvi}

Let us investigate a little more closely iterative projection to multivariate Gaussian PDFs, given their importance in statistics and estimation theory.  

\subsection{Projections}

As mentioned at the end of the last section, we do not have to maintain the same basis from step to step as long as each basis spans the same subspace.  This is a particularly useful maneuver when using the subspace $\mbc G$ of indefinite Gaussians, which are discussed in detail for the multivariate case in Appendix~\ref{sec:indef_gauss}.  Denote the mean and variance of $q^{(i)} \in \mbc G$ as $\mbs\mu^{(i)}$ and $\mbf\Sigma^{(i)}$ and let the basis $\mbf g^{(i)}$ be defined as in (\ref{gaus:1}) and (\ref{gaus:2}).  Note that this basis is orthonormal with respect to $q^{(i)}$.  As such, $\mbf I_{\mbs\alpha}^{(i)} = \ip{\mbf g^{(i)}}{\mbf g^{(i)}}_{q^{(i)}} = \mbf 1$.  Imagine the \ac{PDF} to be approximated is expressed as $p = \norm{\exp(-\phi(\mbf x))} \in \norm{\mbc{B}^2}$.   The coordinates resulting from the next projection are given by (\ref{gaus:7}), namely,
\begin{equation}
\begin{gathered}
    \mbs\alpha_1^{(i+1)} = \ip{\mbf g_1^{(i)}}{p} = {\mbf L^{(i)}}^T
        \mathbb E_{q^{(i)}}\left[\frac{\partial\phi(\mbf x)}{\partial\mbf x^T}\right],
    \\
    \mbs\alpha_2^{(i+1)} = \ip{\mbf g_2^{(i)}}{p} = \sqrt{\mbox{$\frac{1}{2}$}\mbf D^T\mbf D}\,
        \text{vech}\left({\mbf L^{(i)}}^T\mathbb E_{q^{(i)}}\left[\frac{\partial^2\phi(\mbf x)}{\partial\mbf x^T\partial\mbf x}\right]{\mbf L^{(i)}}\right),
\end{gathered}
\end{equation}
where $\mbf L^{(i)}$ issues from the Cholesky decomposition of $\mbf\Sigma^{(i)}$.

The new iteration is
\begin{equation}\label{ipg:1}
    q^{(i+1)} = \underset{(\mbc G, q^{(i)})}{\norm{}\text{proj}} p
        = \norm \exp\left(-{\mbs\alpha_1^{(i+1)}}^T\mbs\gamma_1^{(i)}
            - {\mbs\alpha_1^{(i+1)}}^T\mbs\gamma_1^{(i)}\right).
\end{equation}
Using (\ref{gaus:8}), this becomes
\begin{equation}\label{ipg:1}
    q^{(i+1)} = \norm\exp\biggl(-(\mbf x - \mbs\mu^{(i)})^T\mathbb E_{q^{(i)}}\left[\frac{\partial\phi(\mbf x)}{\partial\mbf x^T}\right]
    -\frac{1}{2}(\mbf x - \mbs\mu^{(i)})^T\mathbb E_{q^{(i)}}\left[\frac{\partial^2\phi(\mbf x)}{\partial\mbf x^T\partial\mbf x}\right]
        (\mbf x - \mbs\mu^{(i)})\biggr),
\end{equation}
which we may cast into the form,
\begin{equation}\label{ipg:1}
    q^{(i+1)} = \norm\exp\left(-\frac{1}{2}(\mbf x - \mbs\mu^{(i+1)})^T{\mbf\Sigma^{(i+1)}}^{-1}
        (\mbf x - \mbs\mu^{(i+1)})\right).
\end{equation}
Herein
\begin{subequations}\label{eq:gviupdate}
\begin{eqnarray}
    {\mbf\Sigma^{(i+1)}}^{-1} &  = & \mathbb E_{q^{(i)}}\left[\frac{\partial^2\phi(\mbf
        x)}{\partial\mbf x^T\partial\mbf x}\right], \\
    {\mbf\Sigma^{(i+1)}}^{-1}\delta\mbs\mu & = & -\mathbb E_{q^{(i)}}\left[\frac{\partial\phi(\mbf
        x)}{\partial\mbf x^T}\right], \\
    \mbs\mu^{(i+1)} & = & \mbs\mu^{(i)} + \delta\mbs\mu
\end{eqnarray}
\end{subequations}
give the updates from $q^{(i)} = \mbc N(\mbs\mu^{(i)},\mbf\Sigma^{(i)})$ to $q^{(i+1)} = \mbc N(\mbs\mu^{(i+1)},\mbf\Sigma^{(i+1)})$ and these are exactly the same as those used in the iterative Gaussian variational inference approach presented by \citet{barfoot_ijrr20}.  We have arrived at the same variational updates but have done so from the framework of a Bayesian Hilbert space, where it becomes abundantly clear that the minimization algorithm is in fact a slightly simplified version of Newton's method.  This also provides the connection back to the classic Gaussian filtering and smoothing algorithms as discussed by \citet{barfoot_ijrr20}.

\subsection{Sparsity in Gaussian Inference}
\label{sec:Gaussiansparsity}

The effect of sparsity as it applies to iterative Gaussian inference is of particular interest.  Let us consider the decomposition of a posterior $p$ in accordance to the general sparsity discussion in Section~\ref{sec:sparsity}; that is,
\begin{equation}
    p = \norm{\exp(-\phi(\mbf{x}))} =  \norm{\exp\left(-\sum_{k=0}^K \phi_k( \mbf{x}_k )\right)} = \norm\bigoplus_{k=0}^K \exp\left(-\phi_k( \mbf{x}_k )\right) = \norm\bigoplus_{k=0}^K p_k,
\end{equation}
where $\phi_k(\mbf{x}_k)$ is the $k$th (negative log) factor expression and $\mbf{x}_k = \mbf P_k\mbf x$.

As in (\ref{gaus:8}), we may express the variational estimate as
\begin{multline}\label{exs:0}
\ifthenelse{\boolean{narrow}}{}{\hfill}
    q^{(i+1)} = \underset{(\mbc{G},q^{(i)})}{\norm{}\mbox{\normalfont proj}} \, p = \norm \exp\biggl( -(\mbf{x} -
        \mbs{\mu}^{(i)})^T\mathbb{E}_{q^{(i)}} \left[ \frac{\partial \phi(\mbf{x})}{\partial \mbf{x}^T}\right] 
        \ifthenelse{\boolean{narrow}}{\biggr. \\ \biggl.}{}
         - \frac{1}{2} (\mbf{x} - \mbs{\mu}^{(i)})^T  \, \mathbb{E}_{q^{(i)}} \left[  \frac{\partial^2 \phi(\mbf{x})}{\partial \mbf{x}^T \partial \mbf{x}} \right]  (\mbf{x} - \mbs{\mu}^{(i)})\biggr),
\ifthenelse{\boolean{narrow}}{}{\hfill}
\end{multline}
using the measure $q^{(i)} = \mathcal{N}(\mbs{\mu}^{(i)}, \mbf\Sigma^{(i)})$.  To take advantage of sparsity, we need to have it reflected in the expectations herein.  The first one leads to
\begin{multline}\label{exs:1}
    \mathbb E_{q^{(i)}}\left[\frac{\partial\phi(\mbf x)}{\partial\mbf x^T}\right]
        = \sum_{k=0}^K \mathbb E_{q^{(i)}}\left[\frac{\partial\phi_k(\mbf x_k)}{\partial\mbf x^T}\right]
        = \sum_{k=0}^K \left(\frac{\partial\mbf x_k}{\partial\mbf x}\right)^T\mathbb E_{q^{(i)}}\left[\frac{\partial\phi_k(\mbf x_k)}{\partial\mbf x_k^T}\right] \\
        = \sum_{k=0}^K \mbf P_k^T\mathbb E_{q^{(i)}}\left[\frac{\partial\phi_k(\mbf x_k)}{\partial\mbf x_k^T}\right] 
        = \sum_{k=0}^K \mbf P_k^T\mathbb E_{q^{(i)}_k}\left[\frac{\partial\phi_k(\mbf x_k)}{\partial\mbf x_k^T}\right],
\end{multline}
given that $\mbf x_k = \mbf P_k\mbf x$.  For each factor $k$, then, we are able to shift the differentiation from $\mbf{x}$ to $\mbf{x}_k$.  We draw attention to the last equality, where the expectation simplifies to using $q^{(i)}_k = q^{(i)}_k(\mbf{x}_k)$, the marginal of the measure for just the variables in factor $k$.  In a similar fashion,
\begin{equation}\label{exs:2}
     \mathbb{E}_{q^{(i)}}\left[ \frac{\partial^2 \phi(\mbf{x})}{\partial \mbf{x}^T \partial \mbf{x}} \right]
        = \sum_{k=0}^K \mbf{P}_k^T \,\mathbb{E}_{q_k^{(i)}} \left[ \frac{\partial^2 \phi_k( \mbf{x}_k )}{\partial \mbf{x}_k^T \partial \mbf{x}_k} \right] \mbf{P}_k
\end{equation}
accounts for the second expectation in (\ref{exs:0}).

The implication of the factorization is that each factor, identified by $\phi_k(\mbf{x}_k)$, is projected onto $\mbc{G}_k$, the marginal subspace associated with variables $\mbf{x}_k$.  The results can then be recombined for the full variational estimate as
\begin{equation}
    q^{(i+1)} = \underset{(\mbc{G},q^{(i)})}{\norm{}\mbox{\normalfont proj}} \, p
        = \norm{}\bigoplus_{k=0}^K \underset{(\mbc{G}_k,q_k^{(i)})}{\mbox{\normalfont proj}} \, p_k = \norm{} \bigoplus_{k=0}^K q_k^{(i+1)}.
\end{equation}
The individual projections of $p_k = \norm{\exp\left(-\phi_k(\mbf{x}_k)\right)}$ onto $(\mbc{G}_k,q^{(i)})$ are
\begin{multline}\label{exs:3}
    q^{(i+1)}_k = \underset{(\mbc{G}_k,q_k^{(i)})}{\norm{}\mbox{\normalfont proj}} \, p_k = \norm \exp\left( -\frac{1}{2} (\mbf{x}_k - \mbs{\mu}_k^{(i+1)} )^T \mbf\Sigma_k^{(i+1)^{-1}} (\mbf{x}_k - \mbs{\mu}_k^{(i+1)} ) \right) \\ = \norm \exp\left( -\frac{1}{2} (\mbf{x} -  \mbf{P}_k^T\mbs{\mu}_k^{(i+1)} )^T  \left( \mbf{P}_k^T \mbf\Sigma_k^{(i+1)^{-1}} \mbf{P}_k \right)(\mbf{x} -  \mbf{P}_k^T \mbs{\mu}_k^{(i+1)} ) \right),
\end{multline}
where
\begin{equation}
    \mbs\mu_k^{(i+1)} = \mbs\mu_k^{(i)} - \mbf\Sigma_k^{(i+1)}\mathbb E_{q_k^{(i)}}\left[\frac{\partial\phi_k(\mbf x_k)}{\partial\mbf x_k^T}\right], \quad
    \mbf\Sigma_k^{(i+1)^{-1}} = \mathbb E_{q_k^{(i)}}\left[\frac{\partial^2 \phi_k( \mbf{x}_k )}{\partial \mbf{x}_k^T \partial \mbf{x}_k}\right].
\end{equation}
It is straightforward to show that the vector sum of $q_k$ from (\ref{exs:3}) reproduces (\ref{exs:0}).  (Note that $\mbf P_k\mbf P_k^T = \mbf 1$ as $\mbf P_k$ is a projection matrix and $\mbf P_k^T$ the corresponding dilation.)  

As explained in detail by \citet{barfoot_ijrr20}, it would be too expensive for practical problems to construct first $\mbf\Sigma^{(i)}$ and then extract the required blocks for the marginals, $q_k^{(i)} = \mathcal{N}(\mbs{\mu}_k^{(i)}, \mbf\Sigma_k^{(i)}) = \mathcal{N}(\mbf{P}_k\mbs{\mu}^{(i)}, \mbf{P}_k\mbf\Sigma^{(i)}\mbf{P}_k^T)$.  We see from the above development that we actually only require the blocks of $\mbf\Sigma^{(i)}$ corresponding to the nonzero blocks of its inverse and the method of \citet{takahashi73} can be used to extract the required blocks efficiently.  \citet{barfoot_ijrr20} provide numerical experiments showing the efficacy of this approach.

%Moreover, if we choose to project onto a larger Hermite basis, we can still continue to use a Gaussian measure since the Hermite basis functions we defined are orthonormal with respect to this measure; this is what was done in the second numerical example above.  This means that even with a larger approximation family, we can continue to exploit the method of \citet{takahashi73} to extract only the required marginals.  

\subsection{Example:  Simultaneous Localization and Mapping}
\label{sec:slam}

A main purpose in the current paper was to show the connection between Bayes space \citep{vandenBoogaart2014} and Gaussian variational inference \citep{barfoot_ijrr20}.  We see that minimizing the KL divergence between a true Bayesian posterior and an approximation can be viewed as iterative projection in Bayes space.  Moreover, by exploiting the sparsity that Bayes space makes clear, the method has the potential to be applied to quite large inference problems.

Following \citet{barfoot_ijrr20}, we consider a batch \ac{SLAM} problem with a robot driving around and building a map of landmarks as depicted in Figure~\ref{fig:robot}.  The robot is equipped with a laser rangefinder and wheel odometers and must estimate its own trajectory and the locations of a number of tubular landmarks. This dataset has been used previously by \citet{barfoot14} to test \ac{SLAM} algorithms.  Groundtruth for both the robot trajectory and landmark positions (this is a unique aspect of this dataset) is provided by a Vicon motion capture system.  The whole dataset is 12,000 timesteps long (approximately 20 minutes of real time).  It was assumed that the data association (i.e., which measurement corresponds to which landmark) is known in this experiment to restrict testing to the state estimation part of the problem.

The state to be estimated is
\begin{equation}
\mbf{x} = \bbm \mbf{x}_0 \\ \mbf{x}_1 \\  \vdots  \\\mbf{x}_K \\ \mbf{m}_1 \\ \vdots \\ \mbf{m}_L \ebm, \quad \mbf{x}_k = \bbm x_k \\ y_k \\ \theta_k \\ \dot{x}_k \\ \dot{y}_k \\ \dot{\theta}_k \ebm, \quad \mbf{m}_\ell = \bbm x_\ell \\ y_\ell \ebm ,
\end{equation}
where $\mbf{x}_k$ is a robot state (position, orientation, velocity, angular velocity) and $\mbf{m}_\ell$ a landmark position.  

For the (linear) prior factor on the robot states we have
\begin{equation}
\varphi_k = \left\{ 
\begin{array}{ll}
\frac{1}{2} (\mbf{x}_0 - \pri{\mbf{x}}_0)^T \pri{\mbf{P}}^{-1} (\mbf{x}_0 - \pri{\mbf{x}}_0)  & k=0 \\
\frac{1}{2} (\mbf{x}_k - \mbf{A} \mbf{x}_{k-1})^T \mbf{Q}^{-1}  (\mbf{x}_k - \mbf{A} \mbf{x}_{k-1})                  & k>0
\end{array}   \right. ,
\end{equation}
with
\begin{gather}
\pri{\mbf{P}} = \mbox{diag}(\sigma_x^2, \sigma_y^2,\sigma_\theta^2, \sigma_{\dot{x}}^2, \sigma_{\dot{y}}^2, \sigma_{\dot{\theta}}^2), \;\; \mbf{A} = \bbm \mbf{1} & T \mbf{1} \\ \mbf{0} & \mbf{1} \ebm, \nonumber \\ \mbf{Q} = \bbm \frac{1}{3} T^3 \mbf{Q}_C  & \frac{1}{2} T^2 \mbf{Q}_C \\ \frac{1}{2} T^2 \mbf{Q}_C & T \mbf{Q}_C \ebm, \quad  \mbf{Q}_C = \mbox{diag}(Q_{C,1}, Q_{C,2},Q_{C,3}),
\end{gather}
where $T$ is the discrete-time sampling period, $Q_{C,i}$ are power spectral densities, and $\sigma_x^2$, $\sigma_y^2$,$\sigma_\theta^2$, $\sigma_{\dot{x}}^2$, $\sigma_{\dot{y}}^2$, $\sigma_{\dot{\theta}}^2$ are variances.  The robot state prior encourages constant velocity \citep[\S 3, p.85]{barfoot17}. 

The (nonlinear) odometry factors, derived from wheel encoder measurements, are 
\begin{equation}
\psi_k = \frac{1}{2} \left( \mbf{v}_k - \mbf{C}_k \mbf{x}_k \right)^T \mbf{S}^{-1}  \left( \mbf{v}_k - \mbf{C}_k \mbf{x}_k \right),
\end{equation}
where
\begin{gather}
\mbf{v}_k = \bbm u_k \\ v_k \\ \omega_k \ebm, \quad \mbf{C}_k = \bbm 0 & 0 & 0 & \cos\theta_k & \sin\theta_k & 0 \\ 0 & 0 & 0 & -\sin\theta_k & \cos\theta_k & 0 \\ 0 & 0 & 0 & 0 & 0 & 1 \ebm, \quad \mbf{S} = \mbox{diag}\left(\sigma^2_u, \sigma_v^2, \sigma_\omega^2 \right).
\end{gather}
The $\mbf{v}_k$ consists of measured forward, lateral, and rotational speeds in the robot frame, derived from wheel encoders; we set $v_k = 0$, which enforces the nonholonomy of the wheels as a soft constraint.  The $\sigma_u^2$, $\sigma_v^2$, and $\sigma_\omega^2$ are measurement noise variances. 

The (nonlinear) bearing measurement factors, derived from a laser rangefinder, are
\begin{equation}
\psi_{\ell,k} = \frac{1}{2} \frac{\left( \beta_{\ell,k} - g(\mbf{m}_{\ell},\mbf{x}_k) \right)^2}{\sigma_r^2},
\end{equation}
with
\begin{equation}
g(\mbf{m}_{\ell},\mbf{x}_k) = \mbox{tan}^{-1}(y_\ell - y_k - d \sin\theta_k,  x_\ell - x_k - d \cos \theta_k) - \theta_k,
\end{equation}
where $\beta_{\ell,k}$ is a bearing measurement from the $k$th robot pose to the $\ell$th landmark, $d$ is the offset of the laser rangefinder from the robot center in the longitudinal direction, and $\sigma_r^2$ is measurement noise variance.  Although the dataset provides range to the landmarks as well, we chose to neglect these measurements to make the problem difficult and therefore show the benefit of taking a variational inference approach.  Our setup is similar to a monocular camera situation, which is known to be a challenging \ac{SLAM} problem.

Putting all the factors together, the posterior that we would like to project to a Gaussian is
\begin{equation}
p(\mbf{x}) = \norm\exp(-\phi(\mbf{x})), \quad \phi(\mbf{x}) =  \sum_{k=0}^K \varphi_k + \sum_{k=0}^K \psi_k + \sum_{k=1}^K \sum_{\ell=1}^L \psi_{\ell,k} + \mbox{constant},
\end{equation}
where it is understood that not all $L = 17$ landmarks are actually seen at each timestep and thus we must remove the factors for unseen landmarks.  

We then carry out iterative projection to a multivariate Gaussian estimate, $q^{(i)} = \mbc N(\mbs\mu^{(i)},\mbf\Sigma^{(i)})$, according to
\begin{equation}
q^{(i+1)} = \underset{(\mbc{G},q^{(i)})}{\norm{}\mbox{\normalfont proj}} \, p,
\end{equation}
where we use~\eqref{eq:gviupdate} at implementation.  

In a sub-sequence of the full dataset comprising $2000$ timestamps with $(x,y,\theta,\dot{x},\dot{y},\dot{\theta})$ for the robot state at each time and $17$ landmarks with $(x,y)$ position, the dimension of the state estimation problem is $N = 12034$.  Spanning the indefinite-Gaussian subspace requires $N$ basis functions for the mean and $\frac{1}{2} N(N+3)$ basis functions for the covariance, or $72,438,663$ basis functions total.  To naively apply the idea of iterative projection to Bayes space would be intractable.  However, by exploiting the sparsity that Bayes space affords (see Sections~\ref{sec:sparsity} and~\ref{sec:Gaussiansparsity}) we are able to do this in a very computationally efficient way.   \citet{barfoot_ijrr20} provide further implementation details of this experiment.

\begin{figure}[t]
\centering
\includegraphics[width=0.6\textwidth]{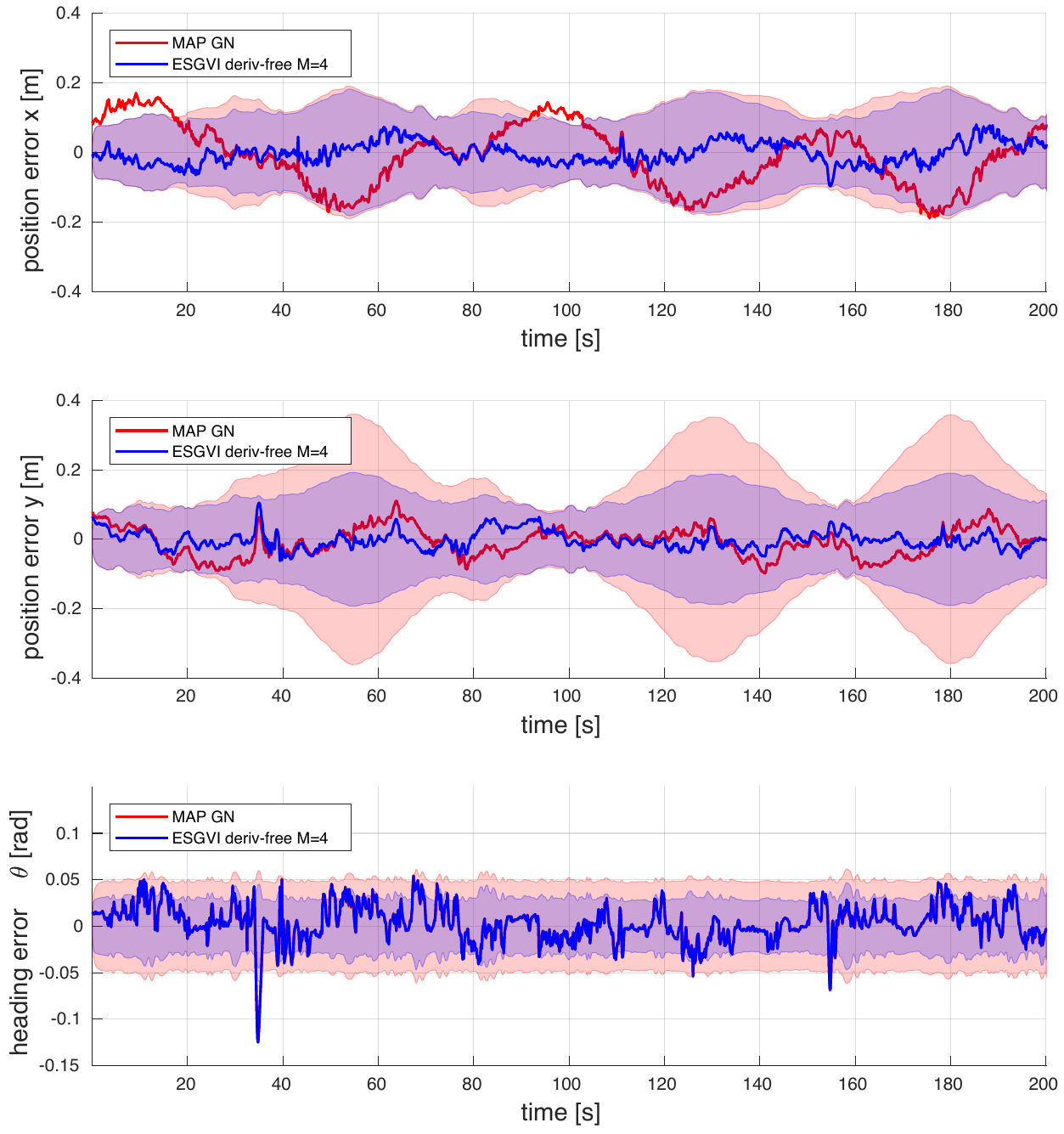}
\caption{Error plots for a portion of the trajectory in the \ac{SLAM} problem conducted by \citet{barfoot_ijrr20} and discussed in Section~\ref{sec:slam}.  The Exactly Sparse Gaussian Variational Inference (ESGVI) algorithm (red) is equivalent to the iterative projection approach described herein.  The Maximum A Posteriori (MAP) Gauss-Newton (GN) algorithm (blue) is the more standard approach to solving this type of problem.  Here we see ESGVI performing slightly better than MAP GN in terms of smaller errors and more consistency (i.e., errors staying within covariance envelope). Note, in the heading error plot, the red mean line is hidden behind the blue one}
\label{fig:exp3}
\end{figure}

Figure~\ref{fig:exp3} shows estimation error plots for the section of the robot's trajectory in Figure~\ref{fig:robot}.  Not only does this show the concepts of Bayes space can be applied to a large problem ($N$ in the range of thousands) but also that there are some situations where it performs slightly better than the standard Maximum A Posteriori (MAP) Gauss-Newton (GN) algorithm.

\section{Discussion}
\label{sec:discussion}

\subsection{Beyond Gaussians}

Much of our discussion has centered on projection to the indefinite-Gaussian subspace and also the use of Gaussian measures in our definition of Bayes space.  This is primarily because we wanted to show the connection between Bayes space and the Gaussian variational inference framework of \citet{barfoot_ijrr20}.  However, we have attempted to lay out the framework to be as general as possible.  

\begin{figure}[t]
\centering
\includegraphics[width=0.7\textwidth]{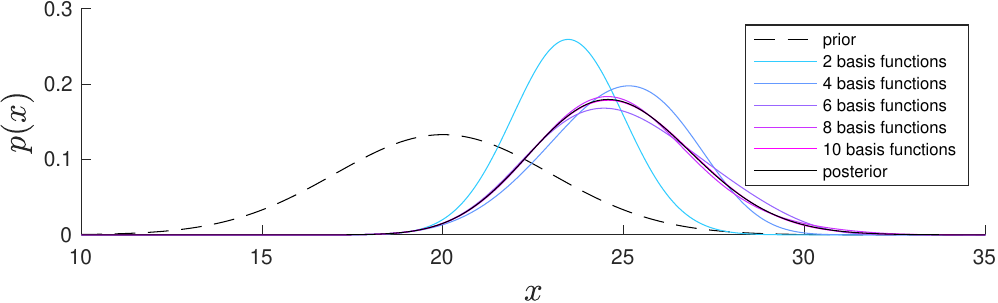}\\ \medskip
\includegraphics[width=0.715\textwidth]{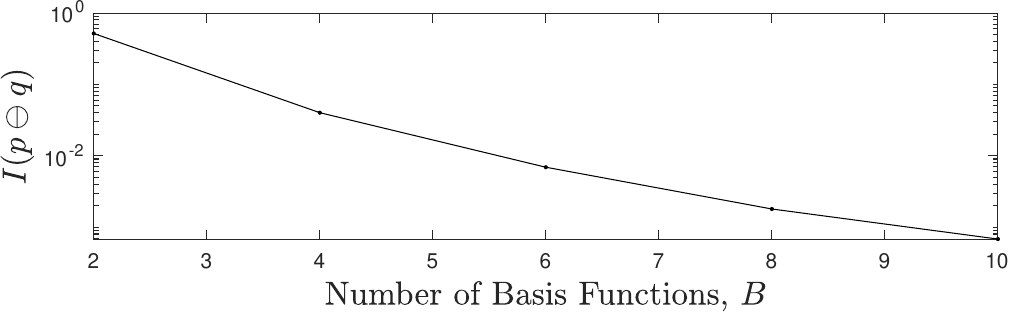}\hspace*{0.00in}
\caption{An example of projection of a posterior onto a finite basis with increasing number of basis functions.  The top panel qualitatively shows that adding more basis functions brings the approximation closer to the posterior.  The bottom shows the same quantitatively where $I(p \ominus q)$ decreases exponentially fast with more basis functions.  The measure was taken to be the prior in this example}
\label{fig:exp4}
\end{figure}

As a teaser of applying the methods beyond Gaussians, we can use $M \geq 2$ Hermite basis functions to see if we can better approximate a \ac{PDF}.  Figure~\ref{fig:exp4} shows that indeed as we project to a higher-dimensional subspace, we are able to better approximate the stereo camera posterior introduced in Section~\ref{subsec:num_ex}.  Here we took the measure $\nu$ to be equal to the prior for the problem.  This shows that even without iteratively updating the measure, we can better approximate the posterior by using more basis functions.  

\begin{figure}[p]
\centering
\includegraphics[width=0.7\textwidth]{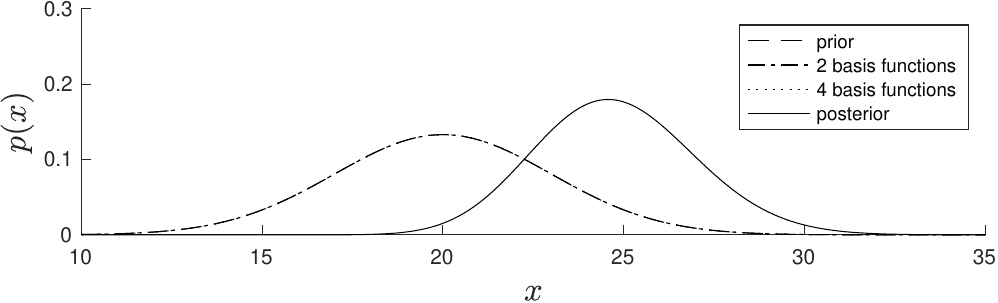}
\includegraphics[width=0.7\textwidth]{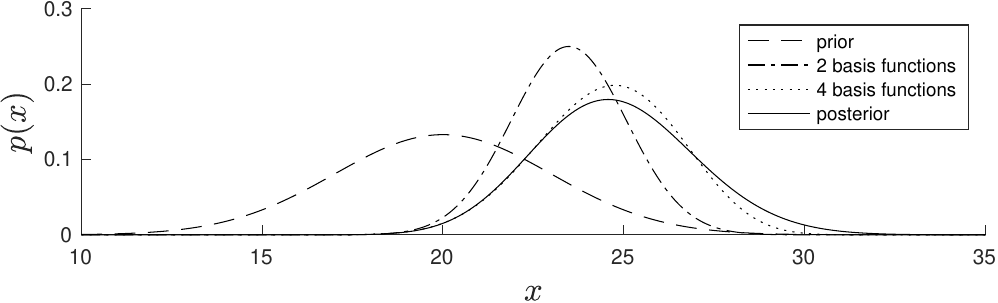}
\includegraphics[width=0.7\textwidth]{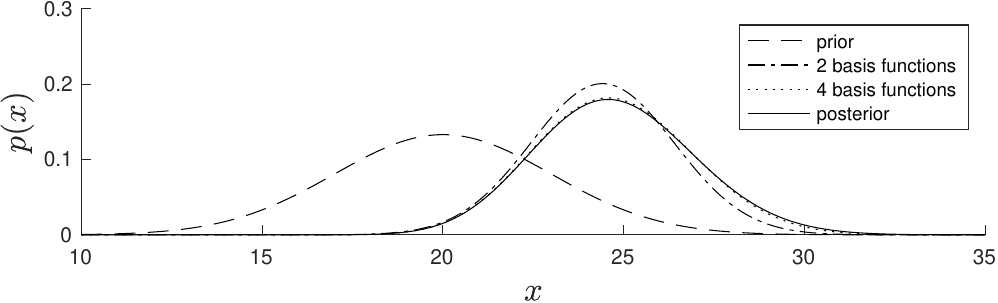}
\includegraphics[width=0.7\textwidth]{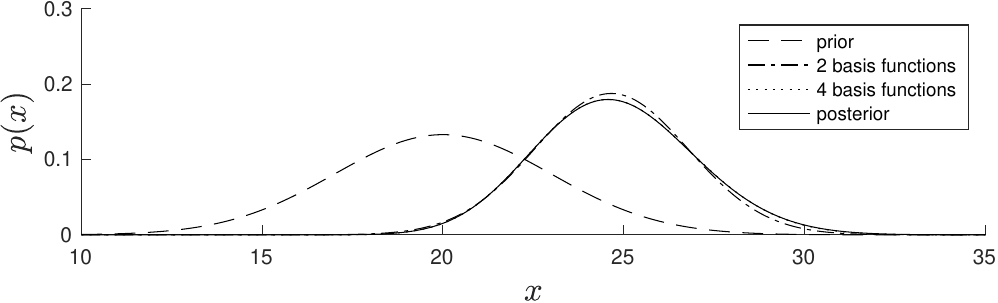}
\includegraphics[width=0.7\textwidth]{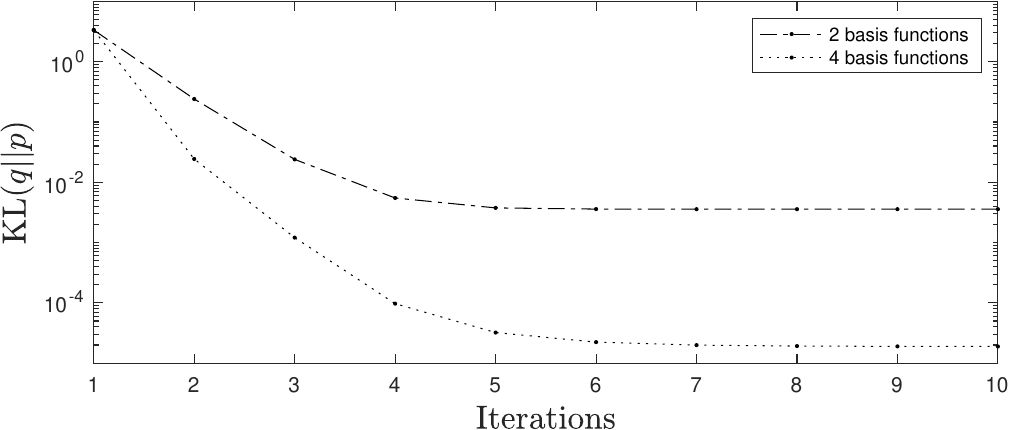}
\caption{Example of iterative projection onto subspaces spanned by $2$ and $4$ Hermite basis functions, where the measure is taken to be the estimate $q^{(i)}$ at the previous iteration (projected to the indefinite-Gaussian subspace) and the basis reorthogonalized at each iteration as described in \S\ref{sec:gvi}.  The estimates were initialized to the prior (first panel) and then iteratively updated (next three panels).  The last panel shows the KL divergence between the estimates and the true posterior for 10 iterations.  We see that the estimate using $4$ basis functions took slightly longer to converge but in the end produced a better approximation of the posterior}
\label{fig:exp5}
\end{figure}

Moreover, we can repeat the iterative projection experiment from Section~\ref{sec:num_ex}, this time with both $2$ and $4$ basis functions for the approximation.  Figure~\ref{fig:exp5} shows the results.  We see that the $4$-basis-function estimate requires a few more iterations to converge than the $2$-basis-function one, but it arrives at a better final approximation as demonstrated by the lower final KL divergence.

\subsection{Limitations and Future Work}

While the results of the previous section make the use of high-dimensional subspaces look promising, there are some limitations still to overcome, which we discuss here.  

First, while the establishment of $\mbc{B}^2$ is mathematically sound, it is actually $\norm{\mbc{B}^2}$ that we are primarily interested in, since we want to approximate valid \acp{PDF} by other valid (simpler) \acp{PDF}.  It seems through our experiments that we have been lucky in the sense that the results of our projections to Bayesian subspaces are valid \acp{PDF}, but there is nothing that actually guarantees this for some of our approximation problems.  For example, consider our one-dimensional Gaussian again:
\begin{equation}
p(x) = \underbrace{\left( -\frac{\mu}{\sigma^2}\right)}_{\alpha_1} \cdot \underbrace{\exp(-x)}_{b_1} \oplus \underbrace{\left(\frac{1}{\sqrt{2}}\frac{1}{\sigma^2}\right)}_{\alpha_2} \cdot \underbrace{\exp\left( -\frac{(x^2 - 1)}{\sqrt{2}} \right)}_{b_2} = \alpha_1 \cdot b_1 \oplus \alpha_2 \cdot b_2,
\end{equation}
which is a member of $\norm{\mbc{G}}$ when $\sigma^2 > 0$.  If we project this vector onto $\mbox{span}\left\{ b_1 \right\}$, just the first Hermite basis vector, the result is
\begin{equation}
\underset{\mbox{\scriptsize span}\left\{ b_1 \right\}}{\mbox{\normalfont proj}} \, p = \alpha_1 \cdot b_1,
\end{equation}
which is no longer a member of $\norm{\mbc{G}}$ since it cannot be normalized to become a valid \ac{PDF}.  Extrapolating from this simple example, it means that truncating a Fourier series at some arbitrary number of terms does not guarantee that the result will be a valid \ac{PDF}.  If we want to extend the Gaussian results to higher-dimensional subspaces, we need to better understand this issue.

Second, even in the case of projecting to the indefinite-Gaussian subspace, guaranteeing that the result is in $\norm{\mbc{G}}$ is quite restrictive.  If the \ac{PDF} to be projected is 
\begin{equation}
p(\mbf{x}) = c \exp( -\phi(\mbf{x})),
\end{equation}
we saw that the projection to the indefinite-Gaussian subspace (see Appendix~\ref{sec:gaussproject}) has the form
\begin{equation}
    \underset{(\mbc G, \nu)}{\text{proj}}\, p = \exp\biggl(-(\mbf x - \mbs\mu)^T\mathbb
        E_\nu\left[\frac{\partial\phi(\mbf x)}{\partial\mbf x^T}\right]
        - \frac{1}{2}(\mbf x - \mbs\mu)^T \underbrace{\mathbb E_\nu\left[\frac{\partial^2\phi(\mbf x)}{\partial\mbf x^T\partial\mbf x}\right]}_{\mbs{\Sigma}^{-1}}(\mbf x - \mbs\mu)\biggr),
\end{equation}
where we have indicated the resulting inverse covariance is $\mbs{\Sigma}^{-1} = \mathbb E_\nu\left[\frac{\partial^2\phi(\mbf x)}{\partial\mbf x^T\partial\mbf x}\right]$.  To guarantee $\mbs{\Sigma}^{-1} > 0$ which would make this a valid \ac{PDF} for any choice of the measure $\nu$, we require that $\phi(\mbf{x})$ is a {\em convex} function of $\mbf{x}$.  This is clearly too restrictive for most real estimation problems involving nonlinear measurement models.  If $\phi(\mbf{x})$ is locally convex, it suggests the measure $\nu$ must be chosen so that its probability mass coincides with this region of local convexity.  This perhaps emphasizes the need to iteratively update the measure in our proposed projection scheme.  However, when the Bayesian posterior and prior are far apart, there is work to be done to understand how best to initialize the measure to ensure the projections wind up in $\norm{\mbc{B}^2}$ in the general setup. 

Finally, in the general case of projecting to a high-dimensional subspace, the measure itself could also be something other than a Gaussian, depending on the basis that is established.  How to carry out the expectations in a computationally efficient and stable way in this case is again future work.  The Hermite basis is cooperative in that the basis functions are orthonormal with respect to a Gaussian measure.  In the high-dimensional \ac{SLAM} problem that we discussed in Section~\ref{sec:slam}, we (i) exploited sparsity inherent in the problem to require only taking expectations over marginals for each measurement factor, and (ii) were also able to exploit the fact that the measure was Gaussian in order to use Gaussian cubature to carry out the expectations somewhat efficiently \citep{barfoot_ijrr20}.  Perhaps there are other bases that could be used for certain problems that admit similar computational conveniences.

It is worth noting that many of the challenges in working with Bayes space stem from the fact that we are attempting to work on infinite domains.  If our interest lies with practical robotic state estimation, Bayes space defined over a finite domain \citep{egozcue06} may be both mathematically simpler as well as more realistic from a practical point of view.  This would of course mean giving up on using Gaussians, for example, which could be replaced by a truncated alternative.

%%%%%%%%%%%%%%%%
\section{Concluding Remarks}

Our principal goal in this work has been to provide a new perspective on the problem of variational inference.  This new vantage point is afforded by considering probability density functions as elements in a Bayesian Hilbert space, where vector addition is a multiplication (perturbation) that accounts for Bayes' rule and scalar multiplication is a exponentiation (powering).  Gaussians and, more generally, exponential families, which are often used in variational inference, are associated with subspaces.  We thus have at our disposal all the familiar instruments of linear algebra.

The use of the Kullback-Leibler divergence $\mbox{KL}(q\|p)$ in variational inference to find the best approximation $q$ to a given posterior $p$ is widespread.  In most approaches, the canvas on which the minimization is carried out is a set, usually convex, or a manifold of admissible functions \citep{Csiszar1975, Csiszar&Tusnady1984, Amari1995a, Adamcik2014, Amari2016}.  `Projections' of $p$ onto the set or manifold are {\em ipso facto} the \ac{PDF} $q$ that minimizes the divergence.  However, in Bayesian space, we may interpret projections as standard linear-algebraic projections, reminding us of a Euclidean world. 

We take particular note of the information geometry of Csisz\'ar and Amari.  They along with their colleagues \citep{Csiszar&Tusnady1984, Amari&al1992} separately developed the {\em em} algorithm---not to be confused with the EM (expectation-maximization) algorithm although the two are in many cases equivalent---to solve the generalized variational problem, which involves a dual minimization of $q$ over its manifold and $p$ over its own.  (The minimum is therefore the minimum `distance' between manifolds.)  The {\em e}-step of the algorithm is performed by making the manifold `flat,' i.e., linear, as a result of using an exponential family of densities.  This flattening is equivalent to thinking in terms of a Bayesian space as we have done here.  Indeed, as we have shown, the natural-gradient-descent algorithm of \citet{amari98} can be explained using this framework as a Newton-like iterative projection.

Based on the inner product of our Bayesian space, we have employed an information measure.  It is proportional to the squared norm of a probability distribution, which can be used to establish a (symmetric and quadratic) divergence between two \acp{PDF}.  The connection to the \ac{KL} divergence is worthwhile mentioning.  Each step in the iterative-projection algorithm presented here for variational inference based on the \ac{KL} divergence amounts to a local minimization of our Bayesian-space divergence.  Admittedly, our iterative-projection approach has some limitations and open issues to resolve.  Particularly, we as yet cannot guarantee that the result of a projection will be a valid PDF and hence starting far from the final posterior estimate can pose challenges.  We hope others may pick up where we have left off to further develop these ideas and overcome current limitations.

The linear structure of Bayes space furthermore allows us to treat sparsity in measurement data very neatly as the vector sum of the measurements, each of which can be expressed as an element in a subspace restricted to the local variables dictated by the sparsity of the problem, for example, as in the simultaneous-localization-and-mapping (SLAM) problem in robotics \citep{barfoot_ijrr20}.  The mean-field approximation in variational inference can be handled in much the same way in this framework.  The factorization of a distribution with respect to a desired family of distributions would again be rendered as a vector sum of \acp{PDF}.

In his fictional autobiography, {\em Zen and the Art of Motorcycle Maintenance}, Robert M. Pirsig notes that ``One geometry cannot be more true than another; it can only be more convenient.''  The same can be said of algebra.  Whether one takes a geometric or algebraic tack in analyzing a problem, it can be agreed that different perspectives offer different views and given a particular problem or even a particular class of problem one tack may sometimes be more convenient than others.  We hope the perspective presented here on variational inference using a Bayesian Hilbert space offers not only convenience in some respects but insight and a degree of elegance as well.

\bibliographystyle{asrl}      
\bibliography{refs.bib}   

\newpage
\appendix
%!TEX root =  robotica.tex
\section{Kronecker Product, vec and vech Operators, and Duplication Matrices}
\label{sec:kron}
For the benefit of the reader, we summarize several identities, which will be used in subsequent appendices, involving the {\em Kronecker product} $\otimes$ and the {\em vectorization} operator $\vec(\cdot)$ that stacks the columns of a matrix:
\begin{equation}\label{kron:1}
\begin{aligned}
\vec(\mbf{a}) & \equiv  \mbf{a} \\
\vec(\mbf{a}\mbf{b}^T) & \equiv  \mbf{b} \otimes \mbf{a} \\
\vec(\mbf{A}\mbf{B}\mbf{C}) & \equiv  (\mbf{C}^T \otimes \mbf{A} )\, \vec(\mbf{B}) \\
\vec(\mbf{A})^T \vec(\mbf{B}) & \equiv  \tr(\mbf{A}^T\mbf{B}) \\
(\mbf{A} \otimes \mbf{B})(\mbf{C} \otimes \mbf{D}) & \equiv  (\mbf{A}\mbf{C}) \otimes (\mbf{B}\mbf{D}) \\
(\mbf{A} \otimes \mbf{B})^{-1} & \equiv  \mbf{A}^{-1} \otimes \mbf{B}^{-1} \\
 (\mbf{A} \otimes \mbf{B})^{T} & \equiv  \mbf{A}^{T} \otimes \mbf{B}^{T}. % \\
\end{aligned}
\end{equation}
It is worth noting that $\otimes$ and $\vec(\cdot)$ are linear operators.

As we will be working with (symmetric) covariance matrices when discussing Gaussians, we would like to be able to represent them parsimoniously in terms of only their unique variables.
Following \citet[\S 18]{magnus19}, we introduce the {\em half-vectorization} operator $\vech(\cdot)$ that stacks up the elements in a column matrix, excluding all the elements above the main diagonal.  The {\em duplication matrix} $\mbf{D}$ allows us to recover a full symmetric matrix from its unique parts:
\begin{equation}
\vec(\mbf{A}) = \mbf{D} \, \vech(\mbf{A})  \qquad \mbox{(symmetric $\mbf{A}$)}.
\end{equation}
It is helpful to consider a simple $2 \times 2$ example:
\begin{equation}
\mbf{A} = \bbm a & b \\ b & c \ebm, \quad \vec(\mbf{A}) = \bbm a \\ b \\ b \\ c \ebm, \quad \mbf{D} = \bbm 1 & 0 & 0 \\ 0 & 1 & 0 \\ 0 & 1 & 0 \\ 0 & 0 & 1 \ebm , \quad
\vech(\mbf{A}) = \bbm a \\ b \\ c \ebm.
\end{equation}
%If we want to convert back to a matrix it is useful to define a corresponding $\matf(\cdot)$ operator\footnote{The `f' indicates we are converting a half vector back into a `full' symmetric matrix.} so that
%\begin{equation}
%\matf(\vech(\mbf{A})) = \mat(\mbf{D} \, \vech(\mbf{A})) = \mat( \vec(\mbf{A}) ) = \mbf{A} \qquad \mbox{(symmetric $\mbf{A}$)}.
%\end{equation}
The {\em Moore-Penrose pseudoinverse} of $\mbf{D}$ will be denoted $\mbf{D}^\dagger$ and is given by
\begin{equation}
\mbf{D}^\dagger = \left( \mbf{D}^T \mbf{D} \right)^{-1} \mbf{D}^T.
\end{equation}
We can then use $\mbf{D}^\dagger$ to convert the vectorization of a matrix into its half-vectorization:
\begin{equation}
\vech(\mbf{A}) = \mbf{D}^\dagger \vec(\mbf{A})  \qquad \mbox{(symmetric $\mbf{A}$)}.
\end{equation}
For our $2 \times 2$ example we have
\begin{equation}
\mbf{D}^\dagger = \bbm 1 & 0 & 0 & 0 \\ 0 & \frac{1}{2} & \frac{1}{2} & 0 \\ 0 & 0 & 0 & 1 \ebm.
\end{equation}
Useful identities involving $\mbf{D}$ are then
\begin{equation}\label{dup:1}
\begin{aligned}
\mbf{D}^\dagger \mbf{D} & \equiv \mbf{1}  \\
{\mbf{D}^\dagger}^T \mbf{D}^T & \equiv \mbf{D} \mbf{D}^\dagger  \\
\mbf{D} \mbf{D}^\dagger \vec(\mbf{A}) & \equiv \vec(\mbf{A})  \qquad\qquad\!\mbox{(symmetric $\mbf{A}$)} \\
\mbf{D} \mbf{D}^\dagger \left( \mbf{A} \otimes \mbf{A} \right) \mbf{D} & \equiv \left( \mbf{A} \otimes \mbf{A} \right) \mbf{D}  \qquad \mbox{(any $\mbf{A}$)},
\end{aligned}
\end{equation}
which can be found in \citet{magnus80}.

\section{Outer Products}
\label{sec:op}

The {\em outer product} $\Phi:\mbc{B}^2 \rightarrow \mbc{B}^2$  of two vectors $b = b(\mbf x), c = c(\mbf x') \in \mbc{B}^2$, denoted $\Phi(\mbf x,\mbf x') = b(\mbf x)\rangle\langle c(\mbf x')$ or briefly $\Phi = b\rangle\langle c$, is defined by its operation on arbitrary $d = d(\mbf x') \in \mbc{B}^2$ as
\begin{equation}
    \Phi(\mbf x,\mbf x')\circledast d(\mbf x') = b(\mbf x)\rangle\langle c(\mbf x')\circledast d(\mbf x') = b(\mbf x) \cdot \ip{c}{d} = \ip{c}{d} \cdot b(\mbf x).
\end{equation}
Thus, dropping the functional dependence,
\begin{equation}
    \ip{a}{\Phi\circledast d} = \ip{a}{b}\ip{c}{d}
\end{equation}
for arbitrary $a \in \mbc{B}^2$.  More generally,
\begin{equation}\label{op:1}
    \Phi = \bigoplus_{i=1}^M \bigoplus_{j=1}^N \phi_{ij} \cdot b_i\rangle\langle c_j,
\end{equation}
where $b_i, c_j \in \mbc{B}^2$ and $\phi_{ij} \in \mathbb R$, so that
\begin{equation}
    \Phi\circledast d = \bigoplus_{i=1}^M \sum_{j=1}^N \phi_{ij}  \ip{c_j}{d} \cdot b_i
\end{equation}
and
\begin{equation}
    \ip{a}{\Phi\circledast d} = \sum_{i=1}^M \sum_{j=1}^N \phi_{ij}\ip{a}{b_i}\ip{c_j}{d}.
\end{equation}
Defining the matrix $\mbf\Phi = [\phi_{ij}] \in \mathbb R^{M\times N}$ and
\begin{equation}
    \mbf b(\mbf x) = \bbm b_1(\mbf x) \\ b_2(\mbf x) \\ \vdots \\ b_M(\mbf x) \ebm, \quad
    \mbf c(\mbf x) = \bbm c_1(\mbf x) \\ c_2(\mbf x) \\ \vdots \\ c_N(\mbf x) \ebm,
\end{equation}
we may abbreviate (\ref{op:1}) to
\begin{equation}
    \Phi(\mbf x,\mbf x') = \mbf b(\mbf x)\rangle\mbf\Phi\langle\mbf c(\mbf x')
\end{equation}
and hence $\ip{a}{\Phi\circledast d} = \ip{a}{\mbf b}\mbf\Phi\ip{\mbf c}{d}$, where $\ip{a}{\mbf b}$ is interpreted as a row and $\ip{\mbf c}{d}$ as a column.

Given an orthonormal basis $\{b_1,b_2\cdots b_M\}$ for a subspace $\mbc S \subset \mbc{B}^2$,
\begin{equation}
    Q = \bigoplus_{m=1}^M b_m\rangle\langle b_m \equiv \mbf b\rangle\langle \mbf b
\end{equation}
is the {\em kernel} of $\mbc S$ and thus, for any $s \in \mbc S$, $Q\circledast s = s$ \citep{Manton&Amblard_arxiv2015}.  For an nonorthonormal basis,
\begin{equation}
    Q = \bigoplus_{m=1}^M \bigoplus_{n=1}^M \kappa_{mn}  \cdot b_m  \rangle\langle b_n
        \equiv \mbf b\rangle\ip{\mbf b}{\mbf b}^{-1}\langle\mbf b,
\end{equation}
where $\kappa_{mn}$ is the $(m,n)$ entry in $\ip{\mbf b}{\mbf b}^{-1}$.  Notationally, $\cdot \left> \mbf{A} \right< \cdot$ indicates an outer product weighted in the middle by an appropriately sized matrix $\mbf{A}$, which in the above example serves to normalize the basis.  In normal matrix algebra it would be equivalent, for example, to writing $\mbf{a} (\mbf{a}^T\mbf{a})^{-1} \mbf{a}^T$, for some column $\mbf{a}$.

Using the outer product, we can write a projection as
\begin{equation}\label{proj:3}
    q^\star(\mbf x) = Q(\mbf x,\mbf x')\circledast p(\mbf x'),
\end{equation}
where
\begin{equation}\label{proj:4}
    Q(\mbf x,\mbf x') = \mbf b(\mbf x)\rangle \ip{\mbf b}{\mbf b}^{-1} \langle\mbf b(\mbf x')
\end{equation}
is the kernel of $\mbc Q$, which plays a similar role to projection matrix.

\section{Hermite Basis}

\subsection{Basis for $\boldsymbol{\mathbb R}$}
\label{sec:herm_r}

Consider the domain over which members of $\mbc{B}^2$ is defined to be $\mathbb{R}$.  We can use the exponentiated Hermite polynomials as a basis for our infinite-dimensional $\mbc{B}^2$; in fact, they prove to be a natural choice \citep{vandenBoogaart2014}.  In one dimension, the first few {\em probabilist's Hermite polynomials} are
\begin{equation}
    H_1(\xi) = \xi, \quad H_2(\xi) = \xi^2-1, \quad H_3(\xi) = \xi^3-3\xi, \quad H_4(\xi) = \xi^4-6\xi^2+3.
\end{equation}
(We exclude $H_0(\xi) = 1$ as the resulting vector is the zero vector; however, it will need to be introduced when considering the domain $\mathbb R^N$ as explained in Appendix~\ref{sec:herm_rN}.)  Owing to the properties of the Hermite polynomials, namely, that
\begin{equation}
    \int_{-\infty}^\infty H_n(\xi)\nu(\xi)\, d\xi = 0, \quad \int_{-\infty}^\infty H_m(\xi) H_n(\xi) \nu(\xi)\, d\xi = n!\, \delta_{mn}, \quad m,n = 1, 2, 3\ldots,
\end{equation}
where $\nu(\xi) = \mathcal{N}(0,1)$ is the standard normal density, we can construct an orthonormal basis for $\mbc{B}^2$ following \citet{egozcue06}.  Accordingly,
\begin{equation}
    \mathbb{E}_\nu[H_n] = 0, \quad \mathbb{E}_\nu[H_m H_n] = n!\,\delta_{mn}, \quad m,n = 1, 2, 3\ldots
\end{equation}
Our basis functions are
\begin{equation}
    h_n(\xi) =\exp\left(-\eta_n(\xi)\right), \quad \eta_n(\xi) = \frac{1}{\sqrt{n!}} H_n(\xi).
\end{equation}
Orthogonality follows as
\begin{multline}
    \ip{h_m}{h_n} = \mathbb{E}_\nu \left[ \eta_m \eta_n \right] -
        \mathbb{E}_\nu \left[ \eta_m \right] \mathbb{E}_\nu \left[ \eta_n \right]
        \ifthenelse{\boolean{narrow}}{\\}{} 
        = \frac{1}{\sqrt{m!n!}} \int_{-\infty}^\infty  H_m(\xi) H_n(\xi) \frac{1}{\sqrt{2\pi}}\exp\left(-\frac{1}{2} \xi^2\right)\, d\xi = \delta_{mn}.
\end{multline}
An arbitrary member $p$ of $\mbc{B}^2$ can be expanded in terms of this Hermite basis.  However, we first need two lemmata, resting on the recursive definition of Hermite polynomials; these are

\begin{lemma}
\label{thm:hermitestein}
For the standard normal measure, $\nu \sim \mathcal{N}(0,1)$,
\begin{equation}
    \mathbb{E}_\nu \left[ H_{n+1}(\xi) f(\xi) \right] = \mathbb{E}_\nu \left[  H_n(\xi) \frac{\partial f(\xi)}{\partial \xi} \right].
\end{equation}
where $f(\xi)$ is a differentiable function and is such that the expectations exist.
\end{lemma}

\paragraph{Proof}
The $n=0$ case,
\begin{equation}
    \mathbb{E}_\nu \left[ \xi f \right] = \mathbb{E}_\nu \left[  \frac{\partial f}{\partial \xi} \right],
\end{equation}
is immediately true by Stein's lemma \citep{stein81}.  For general case $n$,
\begin{multline}
\mathbb{E}_\nu \left[H_{n+1} f \right] = \mathbb{E}_\nu \left[\left( \xi H_n - \frac{\partial H_n}{\partial \xi} \right) f \right] = \mathbb{E}_\nu \left[\frac{\partial}{\partial \xi} \left(H_n f \right) - \frac{\partial H_n}{\partial \xi} f \right] \\ = \mathbb{E}_\nu \left[\frac{\partial H_n}{\partial \xi} f + H_n \frac{\partial f}{\partial \xi}- \frac{\partial H_n}{\partial \xi} f \right] = \mathbb{E}_\nu \left[H_n \frac{\partial f}{\partial \xi} \right],
\end{multline}
where we have used the recurrence relation,
\begin{equation}
    H_{n+1} = \xi H_n - \frac{\partial H_n}{\partial \xi},
\end{equation}
for the Hermite polynomials. 

$\hfill \blacksquare$

\medskip
\begin{lemma}
\label{thm:hermitestein2}
For the standard normal measure, $\nu \sim \mathcal{N}(0,1)$,
\begin{equation}
    \mathbb{E}_\nu \left[ H_n(\xi) f(\xi) \right] = \mathbb{E}_\nu \left[\frac{\partial^n f(\xi)}{\partial \xi^n} \right].
\end{equation}
where $f(\xi)$ is an $n$-fold differentiable function and is such that the expectations exist.
\end{lemma}

\paragraph{Proof}
Repeatedly applying Lemma~\ref{thm:hermitestein},
\begin{multline}
\ifthenelse{\boolean{narrow}}{}{\hfill}
\mathbb{E}_\nu \left[ H_n f \right] = \mathbb{E}_\nu \left[  H_{n-1} \frac{\partial f}{\partial \xi} \right] = \mathbb{E}_\nu \left[ H_{n-2} \frac{\partial^2 f}{\partial \xi^2} \right] = \cdots = \mathbb{E}_\nu \left[  H_{1} \frac{\partial^{n-1} f}{\partial \xi^{n-1}} \right] 
\ifthenelse{\boolean{narrow}}{\\}{}
= \mathbb{E}_\nu \left[ H_0\frac{\partial^n f}{\partial \xi^n} \right] = \mathbb{E}_\nu \left[\frac{\partial^n f}{\partial \xi^n} \right],
\ifthenelse{\boolean{narrow}}{}{\hfill}
\end{multline}
yields the desired result.

$\hfill \blacksquare$

\bigskip
Now consider any $p \in \mbc{B}^2$ expressed as $p(\xi) = c\exp(-\phi(\xi))$.  The coordinates are given by
\begin{equation}\label{herm:1}
    \alpha_n = \ip{h_n}{p} = \frac{1}{\sqrt{n!}}{\mathbb E}_\nu\left[\frac{\partial^n \phi(\xi)}{\partial \xi^n} \right]
\end{equation}
and hence
\begin{equation}
    p(\xi) = \bigoplus_{n=1}^\infty \alpha_n\cdot h_n(\xi) = \exp\left(-\sum_{n=1}^\infty
        \frac{1}{n!}{\mathbb E}_\nu\left[\frac{\partial^n\phi(\xi)}{\partial \xi^n}\right]H_n(\xi)\right).
\end{equation}

We can account for measures other than the standard normal density, say $\nu \sim \mbc N(\mu,\sigma^2)$, by the well known reparameterization `trick',
\begin{equation}
    x = \mu + \sigma \xi,
\end{equation}
which leads to
\begin{equation}
    p(x) = \bigoplus_{n=1}^\infty \alpha_n \cdot h_n\left(\frac{x-\mu}{\sigma}\right)
        = \exp\left( - \sum_{n=1}^\infty  \frac{\sigma^n}{n!}  \mathbb{E}_\nu \left[\frac{\partial^n \phi(x)}{\partial x^n}   \right] H_n\left(\frac{x-\mu}{\sigma}\right)\right).
\end{equation}
It is instructive to rewrite this expression by replacing $-\phi$ with $\ln p$ giving
\begin{equation}
    p(x) = \exp\left( \sum_{n=1}^\infty  \frac{\sigma^n}{n!}  \mathbb{E}_\nu \left[\frac{\partial^n \ln p(x)}{\partial x^n}   \right] H_n\left(\frac{x-\mu}{\sigma}\right)\right).
\end{equation}
This is a Taylor-like expansion of $p$ pivoting on a given mean $\mu$ and standard deviation $\sigma$.

Any subset of the basis functions $\{h_1, h_2, \ldots\}$ establishes a subspace of $\mbc{B}^2$; however, as far as such subspaces are concerned, it would be natural to choose an $M$-dimensional subspace $\mbc H$ spanned by the first $M$ basis functions.  As the basis is orthonormal, the Gram matrix is $\ip{\mbf h}{\mbf h} = \mbf 1$.

The Hermite functions can also be used to generate a basis for $\mbc{B}^2$ on the domain $\mathbb R^N$, which we detail in the next subsection.

\subsection{Basis for $\boldsymbol{\mathbb R^N}$}
\label{sec:herm_rN}

We can extend the results of the previous subsection to create a Hermite basis for $\mbc{B}^2$ on $\mathbb R^N$.  Let
\begin{equation}
    \mbs\eta(\xi) = \frac{1}{\sqrt{n!}} \bbm H_0(\xi) \\ H_1(\xi) \\ \vdots \\ H_M(\xi) \ebm.
\end{equation}
Note that we have reintroduced $H_0(\xi)$ because the basis will be created by all possible combinatorial $N$-products of these functions, one for each variable in $\mbs \xi \in \mathbb R^N$.  However, we will have to exclude the combination made up of only $H_0$ because once again this function gives the zero vector of $\mbc{B}^2$.  We may express this operation as a Kronecker product, i.e.,
\begin{equation}
    \mbs\eta(\mbs\xi) = \mbf C\left(\mbs\eta(\xi_1)\otimes\mbs\eta(\xi_2)\otimes\cdots \otimes\mbs\eta(\xi_N)\right),
\end{equation}
where $\mbf C = [\,\mbf 0\;\;\mbf 1\,]$ contains zero in the first column followed by the identity matrix; this removes the offending function.  Observe that $\mbf C\mbf C^T = \mbf 1$.  The basis is then
\begin{equation}
    \mbf h(\mbs\xi) = \exp(-\mbs\eta(\mbs\xi)).
\end{equation}
The total number of basis functions is $(M + 1)^N - 1$.

This set of basis functions retains its orthonormality because
\begin{multline}
    \ip{\mbf h(\mbs\xi)}{\mbf h(\mbs\xi)}
        = \mathbb{E}_\nu \left[ \left(\mbf C \mbs{\eta}(\xi_1) \otimes \cdots \otimes \mbs{\eta}(\xi_N) \right) \left(\mbf C \mbs{\eta}(\xi_1) \otimes \cdots \otimes \mbs{\eta}(\xi_N) \right)^T \right] \\ = \mbf C \, \mathbb{E}_\nu \left[ \mbs{\eta}(\xi_1)\mbs{\eta}(\xi_1)^T \otimes \cdots \otimes \mbs{\eta}(\xi_N) \mbs{\eta}(\xi_N)^T \right]\mbf C^T
\end{multline}
by a property of the Kronecker product (Appendix~\ref{sec:kron}).  Now
\begin{multline}
  \mathbb{E}_\nu \left[ \mbs{\eta}(\xi_1)\mbs{\eta}(\xi_1)^T \otimes \cdots \otimes \mbs{\eta}(\xi_N) \mbs{\eta}(\xi_N)^T \right] \\
    = \int_{-\infty}^\infty \mbs{\eta}(\xi_1)\mbs{\eta}(\xi_1)^T \nu(\xi_1)\, d\xi_1  \otimes \cdots \otimes \int_{-\infty}^\infty \mbs{\eta}(\xi_N)\mbs{\eta}(\xi_N)^T \nu(\xi_N)\, d\xi_N
        = \mbf{1}_{(M + 1)^N\times (M + 1)^N},
\end{multline}
wherein each of the integrals expresses the orthonormality of the Hermite functions and results in an $(M + 1)\times(M + 1)$ identity matrix.  Hence
\begin{equation}
    \ip{\mbf h(\mbs\xi)}{\mbf h(\mbs\xi)} = \mbf C\mbf{1}_{(M + 1)^N\times (M + 1)^N}\mbf C^T
        = \mbf 1_{[(M+1)^N-1]\times[(M+1)^N-1]}.
\end{equation}

To determine the coordinates of an arbitrary $p \in \mbc{B}^2$, we shall require the multivariate version of Lemma 2:
\begin{lemma}
\label{thm:hermitestein3}
For the standard normal measure, $\nu \sim \mathcal{N}(\mbf{0},\mbf{1})$,
\begin{equation}
    \mathbb{E}_\nu \left[ H_{n_1}(\xi_1)H_{n_2}(\xi_2)\cdots H_{n_N}(\xi_N) f(\mbs{\xi}) \right] = \mathbb{E}_\nu \left[\frac{\partial^{n_1+n_2 + \cdots + n_N} f(\mbs{\xi})}{\partial \xi_1^{n_1}\partial \xi_2^{n_2}\cdots\partial \xi_N^{n_N}} \right], \quad\quad n_k = 1, 2\cdots M.
\end{equation}
where $f:\mathbb R^N \rightarrow \mathbb R$ is $n_k$-fold differentiable in $\xi_k$ and is such that the expectations exist.
\end{lemma}
The proof relies on the use of Lemma~2 for each individual partial derivative; for example, with respect to the variable $\xi_1$,
\begin{equation}
    \mathbb{E}_\nu \bigl[ H_{n_1}(\xi_1)H_{n_2}(\xi_2)\cdots H_{n_N}(\xi_N) f(\mbs{\xi}) \bigr] = \mathbb{E}_\nu \left[ H_{n_1-1}(\xi_1)H_{n_2}(\xi_2)\cdots H_{n_N}(\xi_N) \frac{\partial f(\mbs{\xi})}{\partial \xi_1} \right].
\end{equation}
The product $H_{n_1}(\xi_2)\cdots H_{n_N}(\xi_N)$ has no dependence on $\xi_1$ and can therefore be treated as a constant.  Doing the same for all the other variables and for the indicated number of times leads to the stated result.

We can streamline the notation by defining
\begin{equation}
    \mbs\partial_\xi = \bbm 1 & \frac{\textstyle\partial}{\textstyle\partial \xi} &  \frac{\textstyle\partial^2}{\textstyle\partial \xi^2}
        & \cdots &  \frac{\textstyle\partial^M}{\textstyle\partial \xi^M} \ebm
\end{equation}
and, as above,
\begin{equation}
    \mbs{\partial}_{\mbs\xi} = \mbf C(\mbs{\partial}_{\xi_1} \otimes \mbs{\partial}_{\xi_2} \otimes \cdots \otimes \mbs{\partial}_{\xi_N}).
\end{equation}
Using the measure $\nu = \mbc N(0,1)$, then,
\begin{equation}
    \mbs\alpha = \ip{\mbf h(\mbs\xi)}{p(\mbs\xi)} = \mathbb E_\nu[\mbs{\partial}_{\mbs\xi}p(\mbs\xi)]
\end{equation}
are the coordinates of $p(\mbs\xi) \in \mbc{B}^2$, truncated to however many basis functions we decide to keep.

%%%%%%%%%%%%%%%%%%%%%%
\section{Multivariate Gaussians}

\subsection{Basis for Multivariate Gaussians}
\label{sec:indef_gauss}

Multivariate Gaussians are quintessentially important to statistics and estimation theory.  Gaussians, as traditionally defined with a positive-definite covariance matrix, do not in themselves form a subspace of $\mbc{B}^2$.  We need to expand the set to include covariance matrices that are sign-indefinite.  Let us accordingly define an $N$-dimensional {\it indefinite-Gaussian \ac{PDF}} as
\begin{equation}
    p(\mbf{x}) = c \exp\left( -\frac{1}{2} (\mbf{x} - \mbs{\mu})^T \mbf\Sigma^{-1} (\mbf{x} - \mbs{\mu}) \right),
\end{equation}
which has mean, $\mbs{\mu}$, and symmetric covariance, $\mbf\Sigma$.  The set of all $N$-dimensional, indefinite Gaussians is
\begin{equation}
    \mbc{G} = \left\{ p(\mbf{x}) = c\exp\left( -\frac{1}{2} (\mbf{x} - \mbs{\mu})^T \mbf\Sigma^{-1} (\mbf{x} - \mbs{\mu}) \right) \, \biggl| \, \mbs{\mu} \in \mathbb{R}^N, \mbf\Sigma \in \mathbb{R}^{N \times N}, 0 < c < \infty \right\}.
\end{equation}
It is easy to show that $\mbc G$ is in fact a subspace of $\mbc{B}^2$ as the zero vector is contained therein ($\mbf\Sigma^{-1} = \mbf O$, allowing that $\mbf\Sigma \rightarrow \infty$) and the set is closed under vector addition and scalar multiplication.

To establish $\mbc{G}$ as a Bayesian Hilbert space, we must have an appropriate measure, $\nu$.  In our case, we choose the measure to also be a Gaussian, $\nu = \mathcal{N} \left(\mbs{\mu}, \mbf\Sigma\right) \in \mbc{G}$.  We may thus declare $\mbc G$ to be a Bayesian Hilbert space for a measure $\nu \in \mbc G$.  We will refer to the set of Gaussian \acp{PDF} with positive-definite covariance, $\mbs{\Sigma} > 0$, as $\norm{\mbc{G}} \subset \mbc{G}$.

Several possibilities exist to parameterize Gaussians \citep{barfoot_arxiv2020}.  There are $\frac{1}{2}N(N + 3)$ unique elements contained in the mean and the symmetric covariance matrix on $\mathbb R^N$; hence the dimension of $\mbc G$ is $\frac{1}{2}N(N + 3)$.  We shall construct our basis on a positive-definite choice of covariance $\mbf\Sigma$ that we can decompose in Cholesky fashion, i.e., $\mbf\Sigma = \mbf L\mbf L^T$.  Now consider
\begin{equation}\label{gaus:1}
    \mbs\gamma_1(\mbf x) = \mbf L^{-1}(\mbf x - \mbs\mu), \quad
    \mbs\gamma_2(\mbf x) = \sqrt{\mbox{$\frac{1}{2}$}\mbf D^T\mbf D}\,\text{vech}\,(\mbf L^{-1}(\mbf x - \mbs\mu)(\mbf x - \mbs\mu)^T\mbf L^{-T}),
\end{equation}
wherein $\text{vech}\,(\cdot)$ is the {\it half-vectorization} of its matrix argument and $\mbf D$ is the associated duplication matrix (see Appendix~\ref{sec:kron}).  Note that $\mbs\gamma_1$ is an $N\times 1$ column and $\mbs\gamma_2$ is an $\frac{1}{2}N(N + 1)\times 1$ column.  With a little abuse of notation, we set the basis functions as
\begin{equation}\label{gaus:2}
    \mbf g(\mbf x) = \bbm \mbf g_1(\mbf x) \\ \mbf g_2(\mbf x) \ebm = \exp\left(-
        \bbm \mbs\gamma_1(\mbf x) \\ \mbs\gamma_2(\mbf x) \ebm\right);
\end{equation}
that is, the exponential is applied elementwise.  We claim that $\mbf g(\mbf x)$ is a basis for $\mbc G$.

It is instructive to show that $\mbf g(\mbf x)$ spans $\mbc G$ as well as serving as the proof that it is a basis.  Consider again the reparameterization `trick' given by
\begin{equation}
    \mbf x = \mbs\mu + \mbf L\mbs\xi
\end{equation}
with $\mbs\xi \sim \mbc N(\mbf 0,\mbf 1)$.  This renders (\ref{gaus:2}) as
\begin{equation}
    \mbf g(\mbs \xi) = \bbm \mbf g_1(\mbs \xi) \\ \mbf g_2(\mbs\xi) \ebm = \exp\left(-
        \bbm \mbs\gamma_1(\mbs\xi) \\ \mbs\gamma_2(\mbs \xi)\ebm\right)
        = \exp\left(-
        \bbm \mbs\xi \\\sqrt{\mbox{$\frac{1}{2}$}\mbf D^T\mbf D}\,\text{vech}\,\mbs\xi\mbs\xi^T \ebm\right).
\end{equation}
A (normalized) linear combination of the basis functions can be written as
\begin{equation}
    p(\mbs\xi) = \norm \exp\left(-\mbs\alpha_1^T\mbs\gamma_1(\mbs \xi) - \mbs\alpha_2^T\mbs\gamma_2(\mbs\xi)\right).
\end{equation}
Now
\begin{equation}\label{gaus:3}
    \mbs\alpha_1^T\mbs\gamma_1 = \mbs\alpha_1^T\mbs\xi.
\end{equation}
Also, we can in general express the second set of coordinates as
\begin{equation}
    \mbs\alpha_2 =\sqrt{\mbox{$\frac{1}{2}$}\mbf D^T\mbf D}\;\text{vech}\,\mbf S
\end{equation}
for some symmetric $\mbf S$ that can easily be reconstructed from $\mbs\alpha_2$.  Hence
\begin{equation}\label{gaus:4}
    \mbs\alpha_2^T\mbs\gamma_2 = \frac{1}{2}(\text{vech}\;\mbf S)^T\mbf D^T\mbf D\text{vech}\,\mbs\xi\mbs\xi^T
        = \frac{1}{2}(\text{vec}\,\mbf S)^T\text{vec}\,\mbs\xi\mbs\xi^T
\end{equation}
given the identities $\text{vech}\,\mbf A = \mbf D^\dagger\text{vec}\,\mbf A$ and $\mbf{DD}^\dagger\text{vec}\,\mbf A = \text{vec}\,\mbf A$, where $\mbf D^\dagger$ is the Moore-Penrose inverse of $\mbf D$ (Appendix~\ref{sec:kron}).  Moreover, the identity $(\text{vec}\,\mbf A)^T\text{vec}\,\mbf B = \text{tr}\,\mbf{AB}$ leads to
%given the identity $\text{vech}\,\mbf A = \mbf D^\dagger\text{vec}\,\mbf A$, where $\mbf D^\dagger$ is the Moore-Penrose inverse of $\mbf D$, as well as the third identity in (\ref{dup:1}).  Moreover, $(\text{vec}\,\mbf A)^T\text{vec}\,\mbf B = \tr\,\mbf{AB} = \tr\,\mbf{BA}$, which is the fourth identity in (\ref{kron:1}), leading to
\begin{equation}\label{gaus:5}
    \mbs\alpha_2^T\mbs\gamma_2 = \frac{1}{2}\tr\left(\mbf S\mbs{\xi\xi}^T\right)
        = \frac{1}{2}\mbs{\xi}^T\mbf S\mbs \xi.
\end{equation}
Then
\begin{multline}\label{gaus:6}
    p(\mbf x) = \norm \exp\left(-\mbs\alpha_1^T\mbs\xi - \frac{1}{2}\mbs\xi^T\mbf S\mbs\xi\right)
        = \norm \exp\left(-\frac{1}{2}\left(\mbs \xi + \mbf{S}^{-1} \mbs\alpha_1)^T\mbf S(\mbs \xi +  \mbf{S}^{-1} \mbs\alpha_1\right)\right)\\
        = \norm \exp\left(-\frac{1}{2}\left(\mbf x - (\mbs\mu - \mbf L  \mbf{S}^{-1}\mbs\alpha_1)\right)^T\mbf L^{-T}\mbf S\mbf L^{-1}\left(\mbf x - (\mbs\mu - \mbf L \mbf{S}^{-1}\mbs\alpha_1)\right)\right).
\end{multline}
This can represent any Gaussian distribution, where the mean is $ \mbs\mu - \mbf L  \mbf{S}^{-1}\mbs\alpha_1$ and the covariance $\mbf{LS}^{-1}\mbf L^T$.  Thus $\mbf g$ spans $\mbc G$.  Furthermore, as the dimension of $\mbc G$ is $\frac{1}{2}N(N + 3)$, the number of functions in $\mbf g$, $\mbf g$ is a basis for $\mbc G$.

This basis is, in addition, orthonormal as can be proven in a straightforward fashion by using the reparameterized form $\mbf g(\mbs\xi)$ and recognizing that the entries in $\mbs\gamma_1(\mbs\xi)$ are $\xi_i$ and those in $\mbs\gamma_2(\mbs\xi)$ are either $\xi_i\xi_j$ ($i\not=j$) or $\xi_i^2/\sqrt{2}$.  Hence, $\ip{\mbf g}{\mbf g} = \mbf 1$.

%As is shown in Appendix \ref{sec:gaussproject},
It can be shown that
\begin{equation}
\begin{gathered}
\label{gaus:7}
    \mbs\alpha_1 = \ip{\mbf g_1}{p} = \mbf L^T\mathbb E_\nu\left[\frac{\partial\phi(\mbf x)}{\partial\mbf x^T}\right],
    \ifthenelse{\boolean{narrow}}{\\}{\quad}
    \mbs\alpha_2 = \ip{\mbf g_2}{p} = \sqrt{\mbox{$\frac{1}{2}$}\mbf D^T\mbf D}\,
        \text{vech}\left(\mbf L^T\mathbb E_\nu\left[\frac{\partial^2\phi(\mbf x)}{\partial\mbf x^T\partial\mbf x}\right]\mbf L\right)
\end{gathered}
\end{equation}
%are the coordinates for any given $p(\mbf x) = \norm{\exp(-\phi(\mbf x))} \in \mbc G$.  
are the coordinates for $p(\mbf x) = \norm{\exp(-\phi(\mbf x))} \in \mbc G$.  Another rendering of (\ref{gaus:6}) is
\begin{equation}\label{gaus:8}
    p(\mbf x) = \norm \exp\left(-(\mbf x - \mbs\mu)^T\mathbb
        E_\nu\left[\frac{\partial\phi(\mbf x)}{\partial\mbf x^T}\right]
        - \frac{1}{2}(\mbf x - \mbs\mu)^T\mathbb E_\nu\left[\frac{\partial^2\phi(\mbf x)}{\partial\mbf x^T\partial\mbf x}\right](\mbf x - \mbs\mu)\right),
\end{equation}
which also expresses the projection of a PDF in $\mbc{B}^2$ onto $\mbc G$.

\subsection{Coordinates of Multivariate Gaussian Projection}
\label{sec:gaussproject}

Let $p(\mbf{x}) = c \exp( -\phi(\mbf{x})) \in \mbc{B}^2$.  Projecting onto $\mbc{G}$, the coordinates associated with basis functions $\mbf{g}_1$ are
\begin{eqnarray}
\mbs{\alpha}_1 & = & \ip{\mbf{g}_1}{p} \nonumber \\
& = & \mathbb{E}_\nu \left[ \mbs{\gamma}_1(\mbf{x}) \phi(\mbf{x}) \right] - \mathbb{E}_\nu \left[ \mbs{\gamma}_1(\mbf{x}) \right] \mathbb{E}_\nu \left[ \phi(\mbf{x}) \right] \nonumber \\
& = & \mathbb{E}_\nu \left[ \mbf{L}^{-1} (\mbf{x} - \mbs{\mu}) \phi(\mbf{x}) \right] - \underbrace{\mathbb{E}_\nu \left[ \mbf{L}^{-1} (\mbf{x} - \mbs{\mu})\right]}_{\mbf{0}} \mathbb{E}_\nu \left[ \phi(\mbf{x}) \right] \\
& = & \mbf{L}^{-1} \mbf\Sigma \, \mathbb{E}_\nu \left[ \frac{\partial \phi(\mbf{x})}{\partial \mbf{x}^T}\right] \nonumber \\
& = & \mbf{L}^{T} \mathbb{E}_\nu \left[ \frac{\partial \phi(\mbf{x})}{\partial \mbf{x}^T}\right], \nonumber
\end{eqnarray}
where we have employed Stein's lemma \citep{stein81} to go from the third line to the fourth.  Taking the inner product of these coefficients with the associated basis functions we have
\begin{equation}
\mbs{\alpha}_1^T \mbs{\gamma}_1(\mbf{x}) =  \mathbb{E}_\nu \left[ \frac{\partial \phi(\mbf{x})}{\partial \mbf{x}^T}\right]^T \mbf{L} \mbf{L}^{-1} (\mbf{x} - \mbs{\mu})  = \mathbb{E}_\nu \left[ \frac{\partial \phi(\mbf{x})}{\partial \mbf{x}^T}\right] ^T (\mbf{x} - \mbs{\mu}).
\end{equation}
The coordinates associated with basis functions $\mbf{g}_2$ are
\begin{eqnarray}
\mbs{\alpha}_2 & = & \ip{\mbf{g}_2}{p} \nonumber \\
& = & \mathbb{E}_\nu \left[ \mbs{\gamma}_2(\mbf{x}) \phi(\mbf{x}) \right] - \mathbb{E}_\nu \left[ \mbs{\gamma}_2(\mbf{x}) \right] \mathbb{E}_\nu \left[ \phi(\mbf{x}) \right] \nonumber \\
& = & \mathbb{E}_\nu \left[\sqrt{\mbox{$\frac{1}{2}$}\mbf{D}^T\mbf{D}} \vech\left( \mbf{L}^{-1} (\mbf{x} - \mbs{\mu})(\mbf{x} - \mbs{\mu})^T \mbf{L}^{-T} \right) \phi(\mbf{x}) \right] \nonumber \\ & & \qquad - \; \mathbb{E}_\nu \left[ \sqrt{\mbox{$\frac{1}{2}$}\mbf{D}^T\mbf{D}} \vech\left( \mbf{L}^{-1} (\mbf{x} - \mbs{\mu})(\mbf{x} - \mbs{\mu})^T \mbf{L}^{-T} \right) \right] \mathbb{E}_\nu \left[ \phi(\mbf{x}) \right]  \\
& = & \sqrt{\mbox{$\frac{1}{2}$}\mbf{D}^T\mbf{D}} \vech\left( \mbf{L}^{-1} \left(\mathbb{E}_\nu \left[ (\mbf{x} - \mbs{\mu})(\mbf{x} - \mbs{\mu})^T \phi(\mbf{x}) \right] - \mbf\Sigma \mathbb{E}_\nu\left[\phi(\mbf{x})\right] \right) \mbf{L}^{-T} \right) \nonumber \\
& = & \sqrt{\mbox{$\frac{1}{2}$}\mbf{D}^T\mbf{D}} \vech\left( \mbf{L}^{-1} \mbf\Sigma \, \mathbb{E}_\nu \left[  \frac{\partial^2 \phi(\mbf{x})}{\partial \mbf{x}^T \partial \mbf{x}} \right]  \mbf\Sigma \mbf{L}^{-T} \right) \nonumber \\
& = &\sqrt{\mbox{$\frac{1}{2}$}\mbf{D}^T\mbf{D}} \vech\left( \mbf{L}^{T} \, \mathbb{E}_\nu \left[  \frac{\partial^2 \phi(\mbf{x})}{\partial \mbf{x}^T \partial \mbf{x}} \right]   \mbf{L} \right),\nonumber
\end{eqnarray}
where we have again used Stein's lemma to go from the fourth line to the fifth, this time a double application.  Taking the inner product of these coefficients with the associated basis functions we have
\begin{eqnarray}
\mbs{\alpha}_2^T \mbs{\gamma}_2(\mbf{x}) & = & \frac{1}{2} \vech\left( \mbf{L}^{T} \, \mathbb{E}_\nu \left[  \frac{\partial^2 \phi(\mbf{x})}{\partial \mbf{x}^T \partial \mbf{x}} \right]   \mbf{L} \right)^T \mbf{D}^T\mbf{D} \, \vech\left( \mbf{L}^{-1} (\mbf{x} - \mbs{\mu})(\mbf{x} - \mbs{\mu})^T \mbf{L}^{-T} \right) \nonumber \\
& = & \frac{1}{2} \vec\left( \mbf{L}^{T} \, \mathbb{E}_\nu \left[  \frac{\partial^2 \phi(\mbf{x})}{\partial \mbf{x}^T \partial \mbf{x}} \right]   \mbf{L} \right)^T\underbrace{\mbf{D}^{\dagger^T} \mbf{D}^T\mbf{D}}_{\mbf{D}} \mbf{D}^\dagger \vec\left( \mbf{L}^{-1} (\mbf{x} - \mbs{\mu})(\mbf{x} - \mbs{\mu})^T \mbf{L}^{-T} \right) \nonumber \\
& = & \frac{1}{2} \vec\left( \mbf{L}^{T} \, \mathbb{E}_\nu \left[  \frac{\partial^2 \phi(\mbf{x})}{\partial \mbf{x}^T \partial \mbf{x}} \right]   \mbf{L} \right)^T \underbrace{\mbf{D} \mbf{D}^\dagger \vec\left( \mbf{L}^{-1} (\mbf{x} - \mbs{\mu})(\mbf{x} - \mbs{\mu})^T \mbf{L}^{-T} \right)}_{\mbox{use~\eqref{dup:1}~third~line}}  \nonumber \\
& = & \frac{1}{2} \vec\left( \mbf{L}^{T} \, \mathbb{E}_\nu \left[  \frac{\partial^2 \phi(\mbf{x})}{\partial \mbf{x}^T \partial \mbf{x}} \right]   \mbf{L} \right)^T \vec\left( \mbf{L}^{-1} (\mbf{x} - \mbs{\mu})(\mbf{x} - \mbs{\mu})^T \mbf{L}^{-T} \right)  \\
& = & \frac{1}{2} \tr\left( \mbf{L}^{T} \, \mathbb{E}_\nu \left[  \frac{\partial^2 \phi(\mbf{x})}{\partial \mbf{x}^T \partial \mbf{x}} \right]   \mbf{L}  \mbf{L}^{-1} (\mbf{x} - \mbs{\mu})(\mbf{x} - \mbs{\mu})^T \mbf{L}^{-T} \right)\nonumber \\
& = & \frac{1}{2} \tr\left( (\mbf{x} - \mbs{\mu})^T \mbf{L}^{-T} \mbf{L}^{T} \, \mathbb{E}_\nu \left[  \frac{\partial^2 \phi(\mbf{x})}{\partial \mbf{x}^T \partial \mbf{x}} \right]   \mbf{L}  \mbf{L}^{-1} (\mbf{x} - \mbs{\mu}) \right) \nonumber \\
& = & \frac{1}{2} (\mbf{x} - \mbs{\mu})^T  \, \mathbb{E}_\nu \left[  \frac{\partial^2 \phi(\mbf{x})}{\partial \mbf{x}^T \partial \mbf{x}} \right]  (\mbf{x} - \mbs{\mu}).\nonumber
\end{eqnarray}
Combining these we have
\begin{equation}
    \underset{(\mbc G, \nu)}{\norm{}\text{proj}}\, p = \norm \exp\left(-(\mbf x - \mbs\mu)^T\mathbb
        E_\nu\left[\frac{\partial\phi(\mbf x)}{\partial\mbf x^T}\right]
        - \frac{1}{2}(\mbf x - \mbs\mu)^T\mathbb E_\nu\left[\frac{\partial^2\phi(\mbf x)}{\partial\mbf x^T\partial\mbf x}\right](\mbf x - \mbs\mu)\right)
\end{equation}
for the projection in terms of its Gaussian basis.

\subsection{Gaussian Information}
\label{sec:gauss_info}

We calculate here the information $I$ contained in a multivariate Gaussian distribution, $g(\mbf{x}) = \mathcal{N}\left( \mbs{\mu}^\prime, \mbf\Sigma^\prime \right) \in \norm{\mbc{G}}$.    We have
\begin{equation}
g(\mbf{x}) = \norm{\exp\left( -\phi(\mbf{x}) \right)}
\end{equation}
with
\begin{equation}
\phi(\mbf{x}) = \frac{1}{2} \left( \mbf{x} - \mbs{\mu}^\prime \right)^T {\mbf\Sigma'}^{-1} \left( \mbf{x} - \mbs{\mu}^\prime \right).
\end{equation}
The measure is taken as $\nu = \mbc N(\mbs\mu,\mbf\Sigma)$.

Using our orthonormal basis for $\mbc G$, the information in $g$ is
\begin{equation}\label{gi:1}
    I(g) = \frac{1}{2}\|g\|^2 = \frac{1}{2}\ip{g}{g} = \frac{1}{2}(\mbs\alpha_1^T\mbs\alpha_1 + \mbs\alpha_2^T\mbs\alpha_2),
\end{equation}
where $\mbs\alpha_1$ and $\mbs\alpha_2$ are the coordinates.  As
\begin{equation}
    \mathbb{E}_\nu \left[ \frac{\partial \phi(\mbf{x})}{\partial \mbf{x}^T}\right]
    = {\mbf\Sigma'}^{-1}(\mbs\mu - \mbs\mu'), \qquad
    \mathbb{E}_\nu \left[ \frac{\partial^2 \phi(\mbf{x})}{\partial \mbf{x}^T\partial\mbf{x}}\right] = {\mbf\Sigma'}^{-1},
\end{equation}
these coordinates are, by (\ref{gaus:7}),
\begin{equation}
    \mbs{\alpha}_1 = \mbf{L}^T {\mbf\Sigma'}^{-1} \left( \mbs{\mu} - \mbs{\mu}^\prime \right), \quad
    \mbs{\alpha}_2 = \sqrt{\mbox{$\frac{1}{2}$}\mbf{D}^T\mbf{D}}\; \mbf{D}^\dagger \vec\left( \mbf{L}^{T} {\mbf\Sigma'}^{-1} \mbf{L} \right).
\end{equation}
Hence, from (\ref{gi:1}),
\begin{equation}\label{gi:2}
    I(g) = \frac{1}{2}\left(\left( \mbs{\mu} - \mbs{\mu}^\prime \right)^T{\mbf\Sigma'}^{-1}
        \mbf\Sigma {\mbf\Sigma'}^{-1} \left( \mbs{\mu} - \mbs{\mu}^\prime \right)
        + \frac{1}{2}\mbox{tr}\, {\mbf\Sigma'}^{-1}  \mbf\Sigma {\mbf\Sigma'}^{-1} \mbf\Sigma \right),
\end{equation}
where the second term is a result of the fourth identity in (\ref{kron:1}) and the third in (\ref{dup:1}).
It will, however, be instructive to rewrite the terms as
\begin{multline}
    \left( \mbs{\mu} - \mbs{\mu}^\prime \right)^T{\mbf\Sigma'}^{-1}
        \mbf\Sigma {\mbf\Sigma'}^{-1} \left( \mbs{\mu} - \mbs{\mu}^\prime \right)
        = \mbs{\mu}^{\prime^T} {\mbf\Sigma'}^{-1} \mbf\Sigma {\mbf\Sigma'}^{-1} \mbs{\mu}^\prime - 2\mbs{\mu}^{\prime^T} {\mbf\Sigma'}^{-1} \mbf\Sigma\left( \mbs{\mu}^T \otimes \mbf{1} \right) \vec\, {\mbf\Sigma'}^{-1} \\
        + \left(\vec\, {\mbf\Sigma'}^{-1}\right)^T\left( \mbs{\mu} \otimes \mbf{1} \right) \mbf\Sigma\left( \mbs{\mu}^T \otimes \mbf{1} \right) \vec\, {\mbf\Sigma'}^{-1}
\end{multline}
\vspace*{-\baselineskip}
\begin{equation}
    \mbox{tr}\, {\mbf\Sigma'}^{-1}  \mbf\Sigma {\mbf\Sigma'}^{-1} \mbf\Sigma
        = \left(\vec\, {\mbf\Sigma'}^{-1} \right)^T  (\mbf\Sigma \otimes \mbf\Sigma) \,\vec \, {\mbf\Sigma'}^{-1},
\end{equation}
with the help of the third and fourth identities in (\ref{kron:1}).  Now (\ref{gi:2}) can be neatly expressed as
\begin{equation}\label{gi:3}
    I(g) = \frac{1}{2} \bbm {\mbf\Sigma'}^{-1} \mbs{\mu}^\prime \\ \vec\, {\mbf\Sigma'}^{-1} \ebm^T
    \bbm \mbf\Sigma & -\mbf\Sigma\left( \mbs{\mu}^T \otimes \mbf{1} \right) \\[2mm]
    -\left( \mbs{\mu} \otimes \mbf{1} \right)\mbf\Sigma\; & \frac{1}{2} \left(\mbf\Sigma \otimes \mbf\Sigma \right) + \left( \mbs{\mu} \otimes \mbf{1} \right) \mbf\Sigma \left( \mbs{\mu}^T \otimes \mbf{1} \right) \ebm
    \bbm  {\mbf\Sigma'}^{-1} \mbs{\mu}^\prime \\ \vec\, {\mbf\Sigma'}^{-1} \ebm.
\end{equation}
This is the information contained in the Gaussian $\mbc N(\mbs\mu',\mbf\Sigma')$ although it is conditioned by the choice of measure $\mbc N(\mbs\mu,\mbf\Sigma)$ used to the define the inner product.  Note that as ${\mbf\Sigma'}^{-1}$ tends to zero, indicating a broadening of the distribution, the information also goes to zero.  The expression (\ref{gi:3}) can also be interpreted as simply writing the information using a different basis associated with the so-called {\em natural parameters} of a Gaussian \citep{barfoot_arxiv2020}.

%%%%%%%%%%%%%%%%%%%%%%
\section{Variational Inference Details}

\subsection{Fisher Information Matrix}
\label{sec:gram}

This section reviews the {\em Fisher information matrix} (FIM) and shows that with respect to the coordinates used in a given subspace it is simply the Gram matrix of the chosen basis.

Let $q(\mbf x|\theta) \in \mbc Q$, a finite-dimensional subspace of $\mbc{B}^2$ with basis $B$, depending on some parameter $\theta$.  The Fisher information on $\theta$ with respect to the measure $\nu$ is defined to be the covariance of the {\em score} \citep{fisher22}, i.e.,
\begin{equation}\label{fish:1}
    I_\theta = \mathbb E_\nu\left[\left(\frac{\partial\ln q}{\partial\theta} - \mathbb
        E_\nu\left[\frac{\partial\ln q}{\partial\theta}\right]\right)^2\right]
        = \mathbb E_\nu\left[\left(\frac{\partial\ln q}{\partial\theta}\right)^2\right]
            - \left(\mathbb E_\nu\left[\frac{\partial\ln q}{\partial\theta}\right]\right)^2.
\end{equation}
While our Fisher information may appear slightly unfamiliar, by taking the measure to be the density $\nu = \norm{q}$ then $\mathbb{E}_q [ \partial \ln q / \partial \theta ] = 0$ and we have the traditional version.  We purposely delay setting $\nu = \norm{q}$ to show the connection to Bayes space.

Take $q$ to be expressed as a normalized linear combination of the basis functions $b_n$, that is,
\begin{equation}\label{fish:2}
    q(\mbf x|\theta) = \norm{}\bigoplus_n \alpha_n(\theta)\cdot b_n.
\end{equation}
The score is
\begin{equation}
    \frac{\partial\ln q}{\partial\theta} = \frac{1}{q}\frac{\partial q}{\partial\theta}.
\end{equation}
As $q = \prod_n b_n^{\alpha_n}/\int \prod_n b_n^{\alpha_n}d\mbf x$,
\begin{multline}\label{fish:3}
    \frac{\partial q}{\partial\theta}
        = \frac{\partial}{\partial\theta}\left(\frac{\prod_n b_n^{\alpha_n}}{\int \prod_n b_n^{\alpha_n}d\mbf x}\right)
        \ifthenelse{\boolean{narrow}}{\\}{}
        = \sum_m \left(\frac{\partial\alpha_m}{\partial\theta}\right)\ln b_m\frac{\prod_n b_n^{\alpha_n}}{\int \prod_n b_n^{\alpha_n}d\mbf x} - \frac{\prod_n b_n^{\alpha_n}}{\int \prod_n b_n^{\alpha_n}d\mbf x}
        \frac{\sum_m (\partial\alpha_m/\partial\theta) \int \ln b_m\prod_n b_n^{\alpha_n}}{\int \prod_n b_n^{\alpha_n}d\mbf x}\\
        = q\sum_m\left(\frac{\partial\alpha_m}{\partial\theta}\right)(\ln b_m - \mathbb E_q[\ln b_m]).
\end{multline}
Hence
\begin{equation}\label{fish:4}
    \frac{\partial\ln q}{\partial\theta} = \sum_m \left(\frac{\partial\alpha_m}{\partial\theta}\right)(\ln b_m - \mathbb E_q[\ln b_m]).
\end{equation}
%As $q = c \prod_n b_n^{\alpha_n}$,
%\begin{equation}\label{fish:3}
%    \frac{\partial q}{\partial\theta}
%        = c \frac{\partial}{\partial\theta}\left(\prod_n b_n^{\alpha_n}\right)
%        = c \sum_m \left(\frac{\partial\alpha_m}{\partial\theta}\right)\ln b_m \prod_n b_n^{\alpha_n} 
%        = q\sum_m\left(\frac{\partial\alpha_m}{\partial\theta}\right) \ln b_m.
%\end{equation}
%Hence
%\begin{equation}\label{fish:4}
%    \frac{\partial\ln q}{\partial\theta} = \sum_m \left(\frac{\partial\alpha_m}{\partial\theta}\right)\ln b_m.
%\end{equation}
%(We note in passing that $\partial\ln q/\partial\theta = \sum_k (\partial\alpha_m/\partial\theta)\,\text{clr}\, b_m$, where $\text{clr}\,b_m$ is the {\em centered log-ratio transformation} \citep{vandenBoogaart2014}.)
Substituting (\ref{fish:4}) into (\ref{fish:1}) produces
\begin{equation}
\label{fish:5}
    I_\theta = \sum_m \sum_n \left(\frac{\partial\alpha_m}{\partial\theta}\right)
    \left(\frac{\partial\alpha_n}{\partial\theta}\right)(\mathbb E_\nu[\ln b_m\ln b_n]
        - \mathbb E_\nu[\ln b_m]\mathbb E_\nu[\ln b_n]) 
               = \left(\frac{\partial\mbs\alpha}{\partial\theta}\right)^T\ip{\mbf b}{\mbf b}\left(\frac{\partial\mbs\alpha}{\partial\theta}\right).
 \end{equation}
The traditional Fisher information uses $\norm{q}$ as the measure and we will indicate that explicitly with a subscript on the inner product, e.g., $\ip{\mbf b}{\mbf b}_q$.  We also note that~\eqref{fish:5} still holds in the event that $q$ is not normalized, owing to the nature of the inner product.

We mention for interest that the stochastic derivative of $q(\mbf x|\theta)$ with respect to $\theta$ is
\begin{equation}
    \frac{\eth q}{\eth\theta} = \bigoplus_m \left(\frac{\partial\alpha_m}{\partial\theta}\right)\cdot b_m(\mbf x)
\end{equation}
and so
\begin{equation}
    I_\theta = \left(\frac{\partial\mbs\alpha}{\partial\theta}\right)^T\ip{\mbf b}{\mbf b}\left(\frac{\partial\mbs\alpha}{\partial\theta}\right)
        = \ip{\bigoplus_m \left(\frac{\partial\alpha_m}{\partial\theta}\right)\cdot b_m}
    {\bigoplus_n \left(\frac{\partial\alpha_n}{\partial\theta}\right)\cdot b_n} = \ip{\frac{\eth q}{\eth\theta}}{\frac{\eth q}{\eth\theta}},
\end{equation}
which makes the inner-product expression of the Fisher information coordinate-free.

For multiple parameters, $\theta_1, \theta_2\ldots \theta_K$, the $(m,n)$ entry in the {\em Fisher information matrix} (\ac{FIM}) is
\begin{multline}
    I_{\mbs\theta,mn} = \mathbb E_\nu\left[\left(\frac{\partial\ln q}{\partial\theta_m} - \mathbb
        E_\nu\left[\frac{\partial\ln q}{\partial\theta_m}\right]\right)\left(\frac{\partial\ln q}{\partial\theta_n} - \mathbb
        E_\nu\left[\frac{\partial\ln q}{\partial\theta_n}\right]\right)\right] \\
        = \mathbb E_\nu\left[\frac{\partial\ln q}{\partial\theta_m}\frac{\partial\ln q}{\partial\theta_n}\right]
            - \mathbb E_\nu\left[\frac{\partial\ln q}{\partial\theta_m}\right]\mathbb E_\nu\left[\frac{\partial\ln q}{\partial\theta_n}\right]
\end{multline}
leading to
\begin{equation}
    \mbf I_{\mbs\theta} = \left(\frac{\partial\mbs\alpha}{\partial\mbs\theta}\right)^T
        \ip{\mbf b}{\mbf b}\left(\frac{\partial\mbs\alpha}{\partial\mbs\theta}\right).
\end{equation}
We shall be particularly interested in the \ac{FIM} with respect to the coordinates for a given basis, that is, when $\mbs\theta = \mbs\alpha$.  In this case, the FIM is simply the Gram matrix,
\begin{equation}\label{fish:6}
    \mbf I_{\mbs\alpha} = \ip{\mbf b}{\mbf b}.
\end{equation}
When $q$ is used as the measure, we shall write $\mbf I_{\mbs\alpha} = \ip{\mbf b}{\mbf b}_q$.

\subsection{Derivation of Equation (\ref{kl:6}): Derivative of the Measure in the Inner Product}
\label{sec:ipderiv}

We consider the inner product
\begin{equation}
    \ip{p}{q}_\nu = \mathbb E_\nu[\ln p\ln q] - \mathbb E_\nu[\ln p]\mathbb E_\nu[\ln q]
\end{equation}
with
\begin{equation}
    \nu = \bigoplus_m \alpha_m\cdot b_m.
\end{equation}
We emphasize that here $p$ and $q$ are held fixed.  The partial derivative with respect to $\alpha_n$ is
\begin{equation}\label{ipd:1}
    \frac{\partial}{\partial\alpha_n}\ip{p}{q}_\nu = \frac{\partial}{\partial\alpha_n}\mathbb E_\nu[\ln p\ln q] - \left(\frac{\partial}{\partial\alpha_n}\mathbb E_\nu[\ln p]\right)\mathbb E_\nu[\ln q]
    - \mathbb E_\nu[\ln p]\left(\frac{\partial}{\partial\alpha_n}\mathbb E_\nu[\ln q]\right).
\end{equation}
In general,
\begin{equation}\label{ipd:2}
    \frac{\partial}{\partial\alpha_n}\mathbb E_\nu[\ln r]
        = \int \frac{\partial\nu}{\partial\alpha_n}\ln rd\mbf x
        = \int \nu\frac{\partial\ln\nu}{\partial\alpha_n}\ln rd\mbf x
        = \mathbb E_\nu\left[\frac{\partial\ln\nu}{\partial\alpha_n}\ln r\right].
\end{equation}
(This quantity may in fact alternatively be written as $\ip{b_n}{r}_\nu$.)  The last two derivatives in (\ref{ipd:1}) are accounted for; as for the first, replacing $\ln r$ with $\ln p \ln q$ above, gives
\begin{equation}
    \frac{\partial}{\partial\alpha_n}\mathbb E_\nu[\ln p\ln q]
        = \mathbb E_\nu\left[\frac{\partial\ln\nu}{\partial\alpha_n}\ln p\ln q\right].
\end{equation}
Thus
\begin{equation}
    \frac{\partial}{\partial\alpha_n}\ip{p}{q}_\nu
        = \mathbb E_\nu\left[\frac{\partial\ln\nu}{\partial\alpha_n}\ln p\ln q\right]
         - \mathbb E_\nu\left[\frac{\partial\ln\nu}{\partial\alpha_n}\ln p\right]\mathbb E_\nu[\ln q] - \mathbb E_\nu[\ln p]\mathbb E_\nu\left[\frac{\partial\ln\nu}{\partial\alpha_n}\ln q\right].
\end{equation}
Now we may rewrite this as
\begin{equation}\label{ipd:3}
    \frac{\partial}{\partial\alpha_n}\ip{p}{q}_\nu
        = \mathbb E_\nu\left[\ln p\ln q^{\partial\ln\nu/\partial\alpha_n}\right]
         - \mathbb E_\nu[\ln p]\mathbb E_\nu\left[\ln q^{\partial\ln\nu/\partial\alpha_n}\right] - \mathbb E_\nu\left[\frac{\partial\ln\nu}{\partial\alpha_n}\ln p\right]\mathbb E_\nu[\ln q].
\end{equation}
We recognize that $q^{\partial\ln\nu/\partial\alpha_n}$ is not a \ac{PDF}; however, the self-normalizing feature of the inner product allows us to write
\begin{equation}
    \mathbb E_\nu\left[\ln p\ln q^{\partial\ln\nu/\partial\alpha_n}\right]
         - \mathbb E_\nu[\ln p]\mathbb E_\nu\left[\ln q^{\partial\ln\nu/\partial\alpha_n}\right]
         = \ip{p}{q^{\partial\ln\nu/\partial\alpha_n}}_\nu
         = \ip{p}{\frac{\partial\ln\nu}{\partial\alpha_n}\cdot q}_\nu.
\end{equation}
For the last term in (\ref{ipd:3}), we use (\ref{fish:4}) yielding
\begin{equation}
    \mathbb E_\nu\left[\frac{\partial\ln\nu}{\partial\alpha_n}\ln p\right]\mathbb E_\nu[\ln q]
        = (\mathbb E_\nu[\ln b_n\ln p] - \mathbb E_\nu[\ln b_n]\mathbb E_\nu[\ln p])\mathbb E_\nu[\ln q] = \mathbb E_\nu[\ln q]\ip{b_n}{p}_\nu.
\end{equation}
Finally then
\begin{equation}\label{ipd:4}
    \frac{\partial}{\partial\alpha_n}\ip{p}{q}_\nu
        = \ip{p}{\frac{\partial\ln\nu}{\partial\alpha_n}\cdot q}_\nu - \mathbb E_\nu[\ln q]\ip{b_n}{p}_\nu.
\end{equation}
As the inner product is symmetric in its arguments, this is also
\begin{equation}
    \frac{\partial}{\partial\alpha_n}\ip{p}{q}_\nu
        = \frac{\partial}{\partial\alpha_n}\ip{q}{p}_\nu
        = \ip{q}{\frac{\partial\ln\nu}{\partial\alpha_n}\cdot p}_\nu - \mathbb E_\nu[\ln p]\ip{b_n}{q}_\nu.
\end{equation}
There is a caveat, however, in that we cannot transfer $\partial \nu/\partial \alpha_n$ as the coefficient of $p$ to that of $q$; this is because the coefficient is a function of the domain variables of the PDFs.  That transformation, though, may be expressed as
\begin{equation}
    \ip{q}{\frac{\partial\ln\nu}{\partial\alpha_n}\cdot p}_\nu
        = \ip{p}{\frac{\partial\ln\nu}{\partial\alpha_n}\cdot q}_\nu + \mathbb E_\nu[\ln p]\ip{b_n}{q}_\nu - \mathbb E_\nu[\ln q]\ip{b_n}{p}_\nu.
\end{equation}
We have used the shorthand $\ip{p}{q}_{\partial\nu/\partial\alpha_n}$ to denote the derivative in (\ref{ipd:4}) as in (\ref{kl:6}).

\end{document}